\documentclass{article}

\usepackage{arxiv}

\usepackage[utf8]{inputenc} 
\usepackage[T1]{fontenc}    
\usepackage{hyperref}       
\usepackage{url}            
\usepackage{booktabs} 
\usepackage{amsmath}
\usepackage{amsfonts}       
\usepackage{nicefrac}       
\usepackage{microtype}      
\usepackage{lipsum}		
\usepackage{graphicx}
\usepackage{natbib}
\usepackage{doi}
\usepackage{algorithm}%
\usepackage{algorithmicx}%
\usepackage{algpseudocode}%
\usepackage{subcaption}
\usepackage{multirow}
\usepackage{arydshln}
\usepackage{nameref}
\usepackage{hyperref}
\usepackage{varwidth}

\title{Leveraging Machine Learning for Early Autism Detection via INDT-ASD Indian Database}

\author{ \href{https://orcid.org/0000-0001-8455-282X}{\includegraphics[scale=0.06]{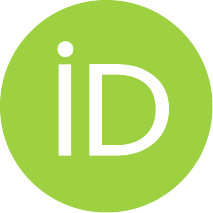}\hspace{1mm}Trapti Shrivastava}\thanks{ \texttt{shri.taps02@gmail.com} Department of Information Technology, Indian Institute of Information Technology, Allahabad, 211013, India} \\
	\And
	\href{https://orcid.org/0009-0001-3485-0666}{\includegraphics[scale=0.06]{orcid.pdf}\hspace{1mm}Harshal Chaudhari} \thanks{Department of Information Technology, Indian Institute of Information Technology, Allahabad, 211013, India} \\
    \And
	\href{https://orcid.org/0000-0002-8818-5673}{\includegraphics[scale=0.06]{orcid.pdf}\hspace{1mm}Vrijendra Singh} \thanks{Department of Information Technology, Indian Institute of Information Technology, Allahabad, 211013, India}
}

\date{}

\renewcommand{\shorttitle}{\textit{arXiv} Template}

\hypersetup{
pdftitle={A template for the arxiv style},
pdfsubject={q-bio.NC, q-bio.QM},
pdfauthor={David S.~Hippocampus, Elias D.~Striatum},
pdfkeywords={First keyword, Second keyword, More},
}

\begin{document}
\maketitle

\begin{abstract}
	Machine learning (ML) has advanced quickly, particularly throughout the area of health care. The diagnosis of neurodevelopment problems using ML is a very important area of healthcare. Autism spectrum disorder (ASD) is one of the developmental disorders that is growing the fastest globally. The clinical screening tests used to identify autistic symptoms are expensive and time-consuming. But now that ML has been advanced, it's feasible to identify autism early on. Previously, many different techniques have been used in investigations. Still, none of them have produced the anticipated outcomes when it comes to the capacity to predict autistic features utilizing a clinically validated Indian ASD database. Therefore, this study aimed to develop a simple, quick, and inexpensive technique for identifying ASD by using ML. Various machine learning classifiers, including Adaboost (AB), Gradient Boost (GB), Decision Tree (DT), Logistic Regression (LR), Random Forest (RF), Gaussian Naive Bayes (GNB), Linear Discriminant Analysis (LDA), Quadratic Discriminant Analysis (QDA), K-Nearest Neighbors (KNN), and Support Vector Machine (SVM), were used to develop the autism prediction model. The proposed method was tested with records from the AIIMS Modified INDT-ASD (AMI) database, which were collected through an application developed by AIIMS in Delhi, India. Feature engineering has been applied to make the proposed solution easier than already available solutions. Using the proposed model, we succeeded in predicting ASD using a minimized set of 20 questions rather than the 28 questions presented in AMI with promising accuracy. In a comparative evaluation, SVM emerged as the superior model among others, with 100 $\pm$ 0.05\% accuracy, higher recall by 5.34\%, and improved accuracy by 2.22\%-6.67\% over RF. We have also introduced a web-based solution supporting both Hindi and English.
\end{abstract}


\section{INTRODUCTION}
Machine learning (ML) is all about bringing together various fields of study, especially as it leads to groundbreaking changes, particularly in the healthcare industry. ML can be used to create tools and processes that can easily and effectively do activities that would typically need human intellect \cite{kusters2020interdisciplinary}. A neuro-developmental disorder known as Autism Spectrum Disorder (ASD) is characterized by recurrent difficulties with speech and nonverbal communication, limited and repetitive behaviors, and social interaction \cite{lord2020autism}. Numerous researchers have examined ASD in children aged eight years. According to research conducted in 2020, 23.1 to 44.9 out of every 1000 eight-year-old children in the US may have ASD \cite{hughes2023prevalence}. Boys were more likely to be affected by autism than girls  \cite{maenner2023prevalence}. Roughly 1 in 8 children in India may have a neurodevelopmental problem, and 1 in 100 children may have ASD \cite{chakrabarti2023autism}. Even with these noteworthy figures, no comprehensive research has been done for ASD on Indian data, the available data is scant.\cite{patankar2022autiscan}.
ASD is characterized by difficulties with communication and social skills, such as difficulty forming and maintaining friendships and interpreting body language. Individuals diagnosed with ASD may also exhibit repetitive habits and struggle with cognitive and learning processes. In other words, individuals could find it difficult to read social cues and communicate with others \cite{lee2023autistic}. ASD sufferers frequently exhibit additional symptoms as well, such as speaking in an odd tone, having trouble understanding and applying language, avoiding eye contact, and having excessive sensitivity to sounds, sights, and other senses. These problems can significantly impact their daily lives, making it difficult for them to manage their jobs, form healthy relationships, and take care of themselves \cite{williams2022definition}. ASD symptoms are often predicted around the age of three and last for the duration of a person's life \cite{ashmawi2022early}. Although the precise origin of ASD is still unknown, evidence indicates that genetics may be a significant factor. It appears that child development, including interactions with surroundings, might be influenced by genetic makeup \cite{ryan2023evidence}. Early detection of ASD is essential for long-term child assistance \cite{shrivastava2022autism}. Studies indicate that autism can be accurately identified by the age of 2, but in practice, most children are not diagnosed until they are between the ages of 3 and 4 \cite{penney2022time}.

Worldwide psychologists refer to various screening tools to observe the behavior of ASD children, such as: 
\begin{itemize}
    \item \textbf{Autism Behaviour Checklist (ABC) \cite{krug1980behavior}} is a questionnaire that was developed in 1980 by Krug and associates that inquires about the behaviors and actions of a child. It's been used for children as young as 3 years old, despite its primary purpose being the detection of autism in school-aged children with severe difficulties. The 57 items on the checklist are divided into five categories: linguistic capabilities, social and self-care abilities, sensory experiences, engagement with others, and how the kid uses their body and objects.

    \item \textbf{Childhood Autism Rating Scale (CARS) \cite{schopler1980toward}} was developed in 1980 to assist in the diagnosis of autism in children as young as two years old by Schopler and colleagues. It was created to match the DSM-III autism criteria, but it subsequently revised its correlation to the DSM-III-R. CARS usually takes twenty to thirty minutes to complete and is intended for direct observation by a qualified practitioner. It evaluates fifteen categories, including the child's capacity to move, imitate, express emotions, interact with objects, adapt to changes, and be sensitive to both sound and vision. It also examines the child's level of activity, mental abilities, sensitivity to touch, anxiety responses, speaking and non-speaking communication abilities, and the clinician's overall assessment of the child.
    
    \item \textbf{Ages and Stages Questionnaires (ASQ) \cite{squires2009ages}} is a set of developmental screening instruments intended to identify possible delays in a child's development from birth to early childhood. It aids in determining whether kids are maturing normally in areas including social skills, problem-solving techniques, physical prowess, and communication. Typically, each ASQ form requires parents to respond to thirty questions. The purpose of these questions is to evaluate many aspects of a child's development.
    The ASQ consists of age-specific questionnaires parents can easily complete and submit at predetermined intervals during their child's early years.

    \item \textbf{Communication and Symbolic Behavior Scales (CSBS) \cite{wetherby1990communication}} was developed to assess infant's and toddler's symbolic behavior and communication abilities, particularly between the ages of 6 and 24 months. Professionals in these fields, like pediatricians and speech-language pathologists, frequently utilize it to identify early indicators of developmental problems. Detailed observations and reporting on the child's communication behaviors involve parents or caregivers in the evaluation process. In the US, where it is heavily utilized, It is crucial in the implementation of early intervention strategies and the monitoring of developmental progress.
    
    \item \textbf{Modified Checklist for Autism in Toddlers (M-CHAT) \cite{robins1999modified}} aims to screen toddlers, especially those aged 16 to 30 months, for ASD. Healthcare practitioners use it extensively to rapidly identify youngsters who might be at risk for autism and need more testing. Parents or other caregivers fill out the checklist by providing answers to questions regarding their child's behavior and developmental milestones. The M-CHAT is a crucial component of early intervention efforts in the US and many other nations, ensuring that kids who may be experiencing developmental delays get the assistance they want as soon as feasible.

    \item \textbf{INCLEN diagnostic tool for autism spectrum disorder (INDT-ASD) \cite{juneja2014inclen}} was created expressly to help India identify children who have ASD. Taking into account the different demographic and healthcare settings in the nation, it is an effort to give a culturally and contextually relevant technique for diagnosing ASD. The integrity, dependability, and applicability of INDT-ASD in the Indian context are guaranteed by its development and validation. To accommodate the crucial early childhood period for ASD diagnosis and intervention, the tool is suitable for children between the ages of 2 and 9.   The parents and caregiver are required to respond to about 38 questions.  To properly diagnose ASD in a diverse population, the emphasis is on obtaining detailed information on a child's behavior and development across various domains.

    \item \textbf{Indian Scale For Assessment of Autism (ISAA) \cite{thenationaltrust2009isaa}} serves as a standardized instrument designed for the diagnosis and assessment of ASD. It is intended to offer a culturally aware framework for detecting ASD in children. The Ministry of Social Justice \& Empowerment established it under its guidance. A wide range of developmental phases, from early childhood through adolescence, are covered by the ISAA, which can be utilized with children ages of 2 to 18. ASD symptoms can be evaluated across a variety of developmental stages and educational levels thanks to this broad age range. Parents and other caregivers play an essential role in the evaluation process. The ISAA relies on parental feedback to a detailed 40-item questionnaire, which yields critical insights into the child's social interactions, communication abilities, repetitive behaviors, and various traits associated with autism.
\end{itemize}

Psychiatrists in India frequently assess ASD patients using the INDT-ASD and ISAA assessment tools \cite{chakrabarti2023autism}. These tools were created specifically for the Indian population and broadly used by certified professionals. The ISAA is a 40-item questionnaire with four anchor points, whereas the INDT-ASD is a 38-item questionnaire with yes/no options. The INDT-ASD questionnaire was revised by \cite{gulati2019development} and introduced as AIIMS-Modified-INCLEN (AMI). In AMI, the number of questions was reduced from 38 to 28. To validate the performance of the AMI, a mobile application was launched and used to gather data from 225 patients. Further, they validated the collected patient's data with DSM-V.

Since ASD cannot be diagnosed through blood tests or similar medical procedures, healthcare professionals depend on an individual's historical and behavioral information for diagnosis. These Screenings are costly and time-consuming. ASD has become a significant priority since early detection can lead to more timely care for people with autism. ML is now used in healthcare, the military, and other industries to assist society.
 
Present methods for ASD screening are often lengthy and significantly dependent on subjective assessments. There is a need to develop an ML-based application to identify ASD for Indian patients. This work aims to provide a method for predicting ASD using ML techniques with the minimal number of questions possible. In addition, we developed a web-based program capable of adequately identifying ASD symptoms using AMI data, both in English and Hindi languages. Our strategy uses the AMI dataset with various ML feature selection algorithms to reduce the required questions. We aim to create precise predictive models through trials with various ML algorithms, intending to pinpoint individuals potentially at risk for ASD swiftly. We intend to evaluate these models using appropriate measures to determine their effectiveness compared to existing methodologies. 


Leveraging the foundational knowledge and methods from prior studies, this research is driven by the exciting possibility of utilizing ML to predict ASD within an Indian data context. The impressive capabilities of ML classifiers showcase the significant potential of ASD classification, emphasizing the urgent need for a solution that is independent of specific hardware and cost-effective. Aiming to overcome the hurdles identified in earlier works while capitalizing on their successes, this study's primary contributions are: 
\begin{itemize}
    \item \textbf{Integration of Ensemble Majority Voting with Feature Selection Techniques:} We innovatively applied an ensemble majority voting system to three prominent feature selection methods: Chi-square (CHS), Recursive Feature Elimination (RFE), and Principal Component Analysis (PCA). This approach is distinct in how it combines the strengths of each method to refine the feature set, reducing the initial set of diagnostic questions from 28 to 20. This reduction not only streamlines the diagnostic process but also mitigates the risk of overfitting, thereby enhancing model generalizability.
    \item \textbf{Comprehensive Evaluation of Machine Learning Classifiers:} We extensively assessed  Logistic Regression (LR), Gaussian Naive Bayes (NB), Decision Tree (DT), Random Forest (RF), Support Vector Machine (SVM), k-nearest neighbors (KNN), Gradient Boosting (GB), AdaBoost (AB), Linear Discriminant Analysis (LDA), and Quadratic Discriminant Analysis (QDA) ML classifiers, utilizing the selected features to train these models comprehensively. Unique to our approach is the meticulous fine-tuning of all possible hyperparameters for each classifier, a process that ensures the optimization of model performance. This exhaustive exploration and optimization of model configurations are rare in existing literature, highlighting our commitment to precision and excellence in model development.
    \item \textbf{Development of a Web-Based ASD Diagnosis Tool:} Leveraging the best-performing ML model identified through rigorous testing and optimization, We have engineered an innovative web-based tool for diagnosing ASD, emphasizing ease of use and broad accessibility. It supports prompt and precise evaluations of ASD in both Hindi and English, catering to diverse user needs. Integrating our optimized ML model into a web-based interface is a significant contribution, as it directly translates complex ML advancements into practical, real-world applications.

\end{itemize}






\section{RELATED WORK} \label{RW}
This section examines some of the most recent research on developing ML models to detect ASD. Nowadays, various ML algorithms, such as conventional ML models, deep neural networks, and transfer learning models, etc. have been explored with different modalities other than screening tools to diagnose and classify ASD according to an autistic individual's behavior, such as eye gaze pattern estimation \cite{alsaidi2024convolutional, rashid2022review, wei2023machine}, functional magnetic resonance imaging (fMRI) \cite{kaur2023review, moridian2022automatic}, gesture analysis \cite{prakash2023video, dia2024video}, and facial feature analysis \cite{uddin2024deep, shrivastava2022autism}. Almost all of these studies are focused on distinguishing and categorizing children with ASD and those who are Typically Developing (TD) using various types of data. These modalities-based solutions have dependencies either on hardware or costly.

To make a hardware-independent and cost-effective solution, a questionnaire-based screen tool is the best option for ASD detection. Popularly earlier, AQ-10-based screen tools have been broadly used by researchers. For Example, Shrivastava, T et al. designed a model using RF on AQ-10 datasets and achieved notable accuracy \cite{shrivastava2024autism}. Uddin M et al. focused on the early diagnosis of ASD in toddlers and developed an AdaBoost-based framework that can effectively analyze the early signs of ASD \cite{uddin2023integrated}. Researchers utilized convolutional neural networks, particle swarm optimization (PSO-CNN), and multilayer perceptron (MLP). They achieved a breakthrough in early ASD diagnosis across various age groups on the AQ-10 datasets \cite{kavitha2023classification, rawat2023asd, hossain2021detecting, jacob2023feature, shinde2023multi}. Leveraging the Support Vector Machine algorithm to analyze AQ-10 test scores and critical characteristics, Artoni et al. efficiently predicted ASD with over 90\% accuracy using just 5 to 7 attributes across three demographic databases \cite{artoni2022autism}. Bisht S and Bisht N developed new supervised machine learning models to predict ASD across different life stages—childhood, adolescence, and adulthood—using a 20-feature dataset from the UCI repository \cite{bisht2022machine}. Farooq et al. applied federated learning to detect ASD using LR and SVM classifiers, achieving 98\% accuracy in children and 81\% in adults, but faced challenges like limited model complexity, data heterogeneity, communication overhead, and lack of transparency \cite{farooq2023detection}. Rashid et al. focused on detecting ASD in children and adults, which achieved reasonable accuracy rates, and clustering methods for scenarios without actual labels \cite{rasul2024evaluation}. Briguglio et al. Employed ML models like RIDGE and LASSO regression on patient data and highlighted how ASD can be distinguished from MSDD using predictors such as gut disturbances and sleep issues \cite{briguglio2023machine}. Hajjej et al. Merged the different ages of ASD datasets and employing SMOTE for data balance, this study utilizes Random Forest and XGBoost classifiers to achieve 94\% accuracy. It was further aimed to customize teaching strategies by analyzing individual performance, underscoring the significance of tailored educational approaches for children with ASD \cite{hajjej2024novel}.

To detect ASD, the researchers collected clinically claimed data other than AQ-10 datasets and performed the ML-based analysis. For example, The Indian Autism Grading Tool (IAGT) has been developed \cite{selvi2023early}. It utilizes the Indian Autism Parental Questionnaire and assigns grades to collected data. The tool has been tested using supervised ML models, with the MLR model achieving the highest accuracy of 97.85\%. The dataset gathered by using IAGT was further utilized to develop a mobile app based on  ML to predict ASD. It outperformed others in sensitivity across different age groups in Tamil Nadu, India. Despite its innovation, the app's low prediction accuracy calls for further enhancements and larger datasets to boost its potential as an early screening tool in healthcare \cite{balasubramanian2024effective}. Chen et al. explored clinical claims of ASD data. They demonstrated how LR and RF models excel at predicting ASD in children, surpassing current tools in early detection with RF showing superior accuracy, highlighting the potential for refining early ASD intervention methods \cite {chen2022early}. Betts et al. presented a novel method for early ASD detection and leveraged ML combined with health administrative data from New South Wales. They have achieved prediction accuracy with an ROC of 0.73. It highlights the potential of using such data for early autism screening to facilitate timely interventions despite needing further refinement for higher accuracy
 \cite{betts2023development}.   
\section{MATERIAL AND METHODOLOGY} \label{MM}

\begin{table}[htbp]
\centering
\caption{AMI Questionnaire Descriptions}
\resizebox{\columnwidth}{!}
{
\begin{tabular}{|l|l|l|} 
\hline
          & Features                     & Description                                                                                                  \\ 
\hline\hline
{[}1]     & Age in months                & Age in months                                                                                                \\ 
\hline
{[}2]     & Gender                       & Male or Female                                                                                               \\ 
\hline
{[}3-10]  & Newla1 to New1a8             & \begin{tabular}[c]{@{}l@{}}Responses regarding Social emotional reciprocity\\type of questions\end{tabular}  \\ 
\hline
{[}11-14] & New1b1 to New1b4             & \begin{tabular}[c]{@{}l@{}}Responses regarding Non verbal communication\\type of questions\end{tabular}      \\ 
\hline
{[}15-17] & New1c1 to New1c3             & Responses on Relationship questions                                                                          \\ 
\hline
{[}18-24] & New2a1 to New2a7             & \begin{tabular}[c]{@{}l@{}}Responses for Stereotyped movement or speech\\questions\end{tabular}              \\ 
\hline
{[}25]    & New2b                        & Responses for Routine questions                                                                              \\ 
\hline
{[}26]    & New2c                        & Responses questions on Fixed interest                                                                        \\ 
\hline
{[}27-30] & New2d1 to New2d4             & Responses based on Sensory symptom questions                                                                 \\ 
\hline
{[}31]    & New Final tool ASD           & -                                                                                                            \\ 
\hline
{[}32]    & New final Summary ASD        & Results as displayed by the Tool used by AIIMS                                                               \\ 
\hline
{[}33]    & DSM5 gold standard diagnosis & Final class displaying ASD or TD                                                                             \\
\hline
\end{tabular}
}
\label{Table1}
\end{table}

\subsection{AIIMS MODIFIED INDT-ASD (AMI) DATASET}

This study used a subset of the comprehensive questionnaire dataset released by Gulati, S. et al. \cite{gulati2019development}. This dataset has 53 features; however, we selected the 33 critical characteristics listed in Table \ref{Table1} for our analysis. Our study is based on data from a group of 225 children (159 boys and 66 girls, median age = 47 months), with 128 diagnosed with ASD and 97 identified as TD using the Gold standard (DSM-5-based expert diagnosis). According to the AIIMS ASD Tool results, 134 of 225 children were diagnosed with ASD and 91 with TD.


\subsection{METHODOLOGY}
\begin{figure*}
    \centering
    \includegraphics[width=\textwidth,height=\textheight,keepaspectratio]{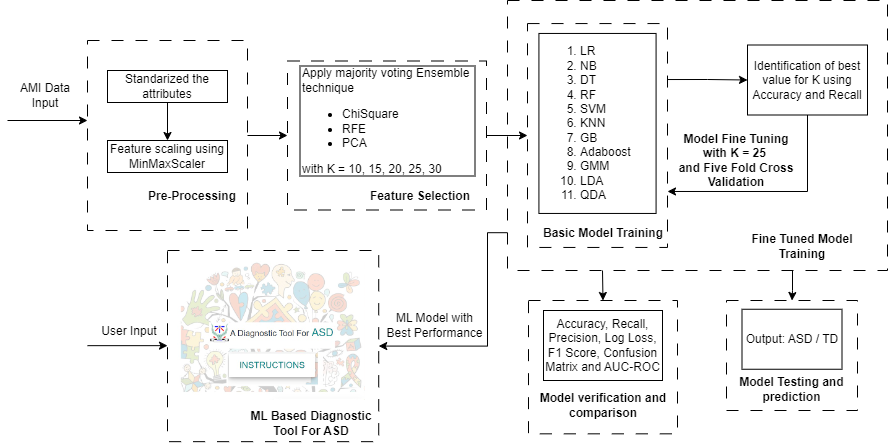}
    \caption{Overview of Proposed Work Flow}
    \label{fig:WF}
\end{figure*}

An overview of our proposed workflow is explained in Figure \ref{fig:WF}. The proposed work is structured into five principal phases, each summarized below:

\textbf{Step 1: Data Pre-Processing}

In our study's preprocessing phase, we refined our dataset by removing non-essential columns like 'Patient ID' and 'New final tool ASD' and selected \textbf{'DSM 5 gold standard diagnosis}' as our target variable. To optimize ML models, we standardized numerical features using the Standard Scaler \cite{scikit-learn} and normalized all features using Min-Max scaling to improve the algorithm's sensitivity to data size. Our preprocessing approach creates a clean, standardized, and structured dataset, providing a solid model creation and analysis platform.

\textbf{Step 2: Feature optimization}

To select the optimized features, we used a multifaceted strategy to determine the most relevant traits for predicting ASD. We use three feature selection techniques in this step: SelectKBest with CHS, RFE with Logistic Regression, and PCA. Further, top K features (with K values of 10, 15, 20, 25, 30) were determined using each selector with the most substantial relationships with the target variable. After selecting the top K feature with individual selectors, we utilized predominant feature selection methods to identify optimal features. Subsequently, the cross-validation approach was employed to determine the most suitable K value for training the machine learning models. It was observed that a K value of 25, along with a set of 20 features, yielded the highest performance across various classifiers.

\textbf{Step 3: Machine Learning Classifiers training for Best Value of K}

In step 3, we assessed various ML classifiers using an 80:20 train-test split and numerous performance metrics. We used cross-validation with a KFold technique to improve our study, improving the classifiers' robust and generalizable performance. This careful evaluation and optimization procedure enabled us to identify the most successful classifiers, establishing the framework for a comparative examination of their strengths and limitations in accomplishing our research objectives.

\textbf{Step 4: Parameter Optimization of ML Classifiers}

We used GridSearchCV to fine-tune all ML classifiers except NB and determined their best parameter combinations. We created comprehensive parameter grids for each model to improve predicted accuracy and used K-fold cross-validation to test various parameter combinations. This detailed process of parameter tuning, critical for maximizing model performance, optimized each classifier with the best set of parameters, improving our models' predictive abilities and providing insights into the impact of various parameters on model performance.

\textbf{Step 5: Model Performance Analysis}

In this step, we assessed our fine-tuned ML models using key metrics such as accuracy, recall, precision, log loss, F1 score, and confusion matrices. Because our research focuses on decreasing missed true positives, we were able to thoroughly evaluate model performance, with a focus on accuracy and recall. The SVM model performed the best, excelling in accuracy and recall, among other measures, making it the ideal model for our study's requirements. This extensive examination demonstrated the SVM's ability to effectively categorize examples while maintaining a high recall, aligning with our research aims.

\section{PROPOSED METHODOLOGY} \label{PW}
\subsection{DATA PRE-PROCESSING}\label{DPP}

Data preprocessing is a crucial phase in training ML models. In this study, we utilize 225 clinically validated datasets. To maintain data integrity, we first analyzed it and found it is relatively balanced data in terms of target class as well as aspects of other attributes, and there is no missing entry in the data. To ensure its compatibility with ML models and maintain the scaling factor, further standardization and normalization techniques have been applied.
\begin{algorithm}
\caption{Feature Standardization}
\begin{algorithmic}[1]

\For{each feature $f$ in the dataset}
    \State Compute the mean $\mu_f$ and standard deviation $\sigma_f$ for feature $f$.
\EndFor

\For{each feature $f$ in the dataset}
    \For{each data point $x_i$ in feature $f$}
        \State Standardize the data point using the formula:
        \[
        z_i = \frac{(x_i - \mu_f)}{\sigma_f}
        \]
        where $x_i$ is the original feature value, $\mu_f$ is the mean, and $\sigma_f$ is the standard deviation of feature $f$.
        \State Apply the standardized value $z_i$ to the data point.
    \EndFor
\EndFor

\State Complete the transformation for all features, standardizing the entire dataset.

\end{algorithmic}
\label{algo:SS}
\end{algorithm}








\subsubsection{Data Standardization}
\mbox{}\\
The standardization technique helps in making the data mean-centric to avoid skewness. Using this strategy, each feature in the dataset will have a mean (\(\mu\)) of 0 and a standard deviation (\(\sigma\)) of 1 as described in Algorithm \ref{algo:SS}. This adjustment is critical for models sensitive to data variation, such as kNN and SVM. The scaler was fitted to our data, converting each feature to have zero mean and unit variance. This approach guarantees that every feature has an equal impact on the outcome, thereby enhancing both the efficiency and accuracy of the algorithm.


    
    

\subsubsection{Data Normalization}
\mbox{}\\
Following standardization, we applied Data Normalization to scale the features to a fixed range – usually [0, 1]. This step is essential for algorithms which react to variations in data scale, such as Gradient Descent, where convergence can be faster if the features are scaled.


    
    

After standardization, we often used data normalization to rescale the features to a given range [0,1]. Scaling features are crucial for algorithms that rely on data scale, such as Gradient Descent, as they speed up convergence. For normalization, we used MinMaxScaler as described in Algorithm \ref{algo:DN}. The scaler changed the data from 0 to 1 while keeping the original distribution and relationships, improving the algorithm's performance and interpretability.

\begin{algorithm}
\caption{Feature Scaling to Range [0, 1]}
\begin{algorithmic}[1]

\For{each feature $f$ in the dataset}
    \State Identify $min_f$ and $max_f$, the minimum and maximum values of $f$.
\EndFor

\For{each feature $f$ in the dataset}
    \For{each data point $x_i$ in feature $f$}
        \State Calculate the normalized value $x_{norm}$ using the formula:
        \State $x_{norm} = \frac{x_i - min_f}{max_f - min_f}$
        \State Apply the calculated $x_{norm}$ to the data point.
    \EndFor
\EndFor

\State Complete the transformation for all features, normalizing the entire dataset to the range [0, 1].

\end{algorithmic}
\label{algo:DN}
\end{algorithm}

\subsection{FEATURE OPTIMIZATION} \label{FS}

Feature selection/optimization is an essential phase for developing ML models. Selecting a subset of relevant features can help improve model interpretability, reduce overfitting, and lower computing costs. For our investigation, we used three different feature selection strategies listed below.

\subsubsection{Chi-Squared Test (CHS)}
\mbox{}\\
The Chi-Square test is a statistical tool for determining if two variables are independent \cite{peters1940chi}. It is used extensively in feature selection for classification challenges. It compares the actual distribution of classes based on feature values to what we would predict if the feature values did not affect the class distribution using \[
\chi^2 = \sum_{i=1}^{n}\sum_{j=1}^{m}\frac{(O_{ij} - E_{ij})^2}{E_{ij}}
\]. where:
\begin{itemize}
    \item $O_{ij}$ represents observed frequency of $x_i$ and $y_j$.
    \item $E_{ij}$ represents expected frequency of $x_i$ and $y_j$ if $X$ and $Y$ are independent.
    \item $n$ is the number of categories in $X$.
    \item $m$ is the number of classes in $Y$.
\end{itemize}

Features with higher $\chi^2$ values are considered more important. Essentially, it assists in identifying features whose values appear to be unrelated to class distribution, making it essential for fine-tuning our feature selection.

\begin{algorithm}
\caption{Algorithm for selecting the optimal feature set using an ensemble majority voting technique}
\begin{algorithmic}[1]
\Function{FS\_CHS}{$X, y, K$}
    \State Apply SelectKBest with chi2 and select top $K$ features.
    \State Return selected features and their scores.
\EndFunction

\Function{FS\_RFE}{$X, y, K$}
    \State Apply RFE with LogisticRegression to select top $K$ features.
    \State Return selected features and their rankings.
\EndFunction

\Function{FS\_PCA}{$X, K$}
    \State Apply PCA and select top $K$ components.
    \State Return selected features based on component loadings.
\EndFunction

\Function{MajorityVoting}{$\text{selectors}$}
    \State Combine feature selection results from all selectors.
    \State Count feature occurrences across selectors.
    \State Select common features appearing in all selectors.
    \State Return final feature set.
\EndFunction

\State $K \gets 10$ \Comment{Number of features to select}

\State $X_{\text{chi2}}, \text{scores}_{\text{chi2}} \gets \Call{FS\_CHS}{X, y, K}$
\State $X_{\text{rfe}}, \text{rankings}_{\text{rfe}} \gets \Call{FS\_RFE}{X, y, K}$
\State $X_{\text{pca}}, \text{components}_{\text{pca}} \gets \Call{FS\_PCA}{X, K}$

\State $X_{\text{majority}} \gets \Call{MajorityVoting}{[X_{\text{chi2}}, X_{\text{rfe}}, X_{\text{pca}}]}$

\Function{SaveFeatures}{$X_{\text{majority}}, y, \text{path}$}
    \State Concatenate $X_{\text{majority}}$ and $y$.
    \State Save concatenated DataFrame to the specified path.
\EndFunction

\State \Call{SaveFeatures}{$X_{\text{majority}}, y, \text{'data.csv'}$}

\end{algorithmic}
\label{algo2}
\end{algorithm}

\subsubsection{Recursive Feature Elimination (RFE)}
\mbox{}\\
RFE \cite{chen2007enhanced}  is a technique for determining which features are most relevant to an ML model. Here is how RFE operates: An ML model is initially trained with all available features. The model then determines the importance of each feature, typically through its internal algorithms designed to rank their significance. After that, it eliminates the least essential feature, retrains the model with the remaining features, and repeats this process, removing one feature at a time until achieving the desired number of features. While RFE's methodology depends on the specific model, its core strategy involves keeping the most impactful features based on the model's evaluation criteria, like coefficients or feature importance scores from logistic regression, support vector machines, or tree-based models.
\subsubsection{Principal Component Analysis (PCA)}
\mbox{}\\
PCA \cite{abdi2010principal} is designed to highlight variability and uncover significant patterns within a dataset. It simplifies the exploration and visualization of data by reducing its dimensions. PCA operates by pinpointing directions that maximize data variance. The technique produces a series of $p$ unit vectors as principal components within a precise coordinate space, with each vector representing the direction of the best-fit line that is orthogonal to the preceding vectors up to the $i$-th order.

The transformation of data to the principal components is given by:
\[
\mathbf{X}_{\text{pca}} = \mathbf{X}\mathbf{W}
\]
Where:
\begin{itemize}
    \item $\mathbf{X}$ is the original data matrix.
    \item $\mathbf{W}$ is the matrix of principal components (eigenvectors of the covariance matrix of $\mathbf{X}$).
\end{itemize}

The amount of variance each principal component captures is reflected in the eigenvalue associated with the eigenvector.
\subsection{PROPOSED FEATURE SELECTION METHODOLOGY BASED ON MAJORITY VOTING} \label{MV}
\mbox{}\\
Majority voting is an ensemble technique used in feature selection to combine the results from multiple methods. The premise is to count the frequency of each feature being selected across different methods and then select those features that are chosen most often. This consensus approach aims to achieve a more reliable and stable selection by considering the intersection of multiple feature ranking criteria. By employing Majority Voting, the feature set is likely to be more predictive and less prone to the biases of individual feature selection algorithms.

As described in algorithm \ref{algo2}, we have used three different selectors, CHS, RFE, and PCA, on different values of K. After getting features from the individual selectors, we have applied majority voting and finalized the minimized set of standard features from all the selectors.

Let $F_1, F_2, ..., F_m$ represent the sets of features chosen through $m$ distinct methods of feature selection, each applied to the same dataset. The aggregated set $F_{majority}$ is defined as:

\begin{align}
F_{majority} &= \{f | f \in F_i \, \text{for} \, i = 1, 2, \ldots, m \nonumber \\
&\quad \text{and} \, \sum_{i=1}^{m} I(f \in F_i) \geq m\}
\end{align}

where $I$ is the indicator function that is 1 if the feature $f$ is in the set $F_i$ selected by the $i$-th method and 0 otherwise. The threshold of $m$ signifies the majority.

\begin{table*}
\centering
\caption{Feature Sets Identified for Different K Values}
\label{tab:my_label}
\resizebox{\columnwidth * 2}{!}
{
\begin{tabular}{|c|c|c|c|c|c|c|c|c|c|c|c|c|c|c|c|} 
\hline
\multirow{2}{*}{Rank} & \multicolumn{3}{c|}{\textbf{K10(4)}}                 & \multicolumn{3}{c|}{\textbf{K15(8)}}                 & \multicolumn{3}{c|}{\textbf{K20(11)}}                & \multicolumn{3}{c|}{\textbf{K25(20)}}                & \multicolumn{3}{c|}{\textbf{K30(30)}}                                      \\ 
\cline{2-16}
                      & CHI             & RFE             & PCA             & CHI             & RFE             & PCA             & CHI             & RFE             & PCA             & CHI             & RFE             & PCA             & CHI                    & RFE                    & PCA                     \\ 
\hline\hline
1                     & \textbf{New1a3} & \textbf{New2a6} & New1a4          & \textbf{New1a3} & \textbf{New2a6} & \textbf{New1a3} & \textbf{New1a3} & \textbf{New2a6} & \textbf{New1a3} & \textbf{New1a3} & \textbf{New2a6} & New1c3          & \textbf{New1a3}        & \textbf{New2a6}        & New1a7                  \\ 
\hline
2                     & \textbf{New2a6} & \textbf{New1a3} & New1a5          & \textbf{New2a6} & \textbf{New1a3} & New1b3          & \textbf{New2a6} & \textbf{New1a3} & New1b3          & \textbf{New2a6} & \textbf{New1a3} & \textbf{New2a3} & \textbf{New2a6}        & \textbf{New1a3}        & New1c3                  \\ 
\hline
3                     & \textbf{New2a7} & \textbf{New2a7} & \textbf{New2b}  & \textbf{New2a7} & \textbf{New2a7} & New1a4          & \textbf{New2a7} & \textbf{New2a7} & \textbf{New2a6} & \textbf{New2a7} & \textbf{New2a7} & \textbf{New1a3} & \textbf{New2a7}        & \textbf{New2a7}        & \textbf{New1a6}         \\ 
\hline
4                     & \textbf{New2b}  & \textbf{New2b}  & New1b3          & \textbf{New2b}  & New2c           & \textbf{New1a5} & \textbf{New2b}  & New2c           & New2a5          & \textbf{New2b}  & \textbf{New2c}  & \textbf{New2a5} & \textbf{New2b}         & \textbf{New2c}         & \textbf{New1a8}         \\ 
\hline
5                     & New2c           & New2d4          & \textbf{New2a7} & New2c           & \textbf{New2b}  & \textbf{New2a6} & New2c           & \textbf{New2b}  & New1a4          & \textbf{New2c}  & \textbf{New2b}  & \textbf{New1b3} & \textbf{New2c}         & \textbf{New2b}         & \textbf{New2a1}         \\ 
\hline
6                     & New2d1          & New2a1          & New2d1          & \textbf{New2d1} & New2d4          & \textbf{New2a7} & \textbf{New2d1} & New2d4          & \textbf{New1a5} & \textbf{New2d1} & \textbf{New2d4} & \textbf{New2a6} & \textbf{New2d1}        & \textbf{New2d4}        & \textbf{New1a2}         \\ 
\hline
7                     & New1c1          & New1a7          & New2a5          & \textbf{New1c1} & \textbf{New2a1} & New1b2          & \textbf{New1c1} & \textbf{New2a1} & \textbf{New2a7} & \textbf{New1c1} & \textbf{New2a1} & \textbf{New2a7} & \textbf{New1c1}        & \textbf{New2a1}        & \textbf{New2a3}         \\ 
\hline
8                     & New2a1          & New1a5          & \textbf{New1a3} & \textbf{New2a1} & \textbf{New2d1} & New2a5          & \textbf{New2a1} & \textbf{New2d1} & New1b2          & \textbf{New2a1} & \textbf{New2d1} & New1a2          & \textbf{New2a1}        & \textbf{New2d1}        & \textbf{New2a5}         \\ 
\hline
9                     & New1a7          & New1b1          & Gender          & New1a7          & New1a7          & \textbf{New2b}  & \textbf{New1a7} & \textbf{New1a7} & \textbf{New2a3} & \textbf{New1a7} & \textbf{New1a7} & \textbf{New1a5} & \textbf{New1a7}        & \textbf{New1a7}        & \textbf{New1c2}         \\ 
\hline
10                    & New1b2          & New1a4          & \textbf{New2a6} & New1b2          & \textbf{New1c1} & \textbf{New1c1} & New1b2          & New2d2          & \textbf{New2b}  & \textbf{New1b2} & \textbf{New2d2} & New1a8          & \textbf{New1b2}        & \textbf{New2a2}        & \textbf{New1a3}         \\ 
\hline
11                    &                 &                 &                 & New2d4          & \textbf{New1a5} & \textbf{New2d1} & New2d4          & \textbf{New1c1} & \textbf{New1c1} & \textbf{New2d4} & \textbf{New1c1} & \textbf{New1a4} & \textbf{New2d4}        & \textbf{New2d2}        & \textbf{New1b3}         \\ 
\hline
12                    &                 &                 &                 & \textbf{New1a5} & New1b1          & Gender          & \textbf{New1a5} & \textbf{New1a5} & \textbf{New1b1} & \textbf{New1a5} & \textbf{New2d3} & \textbf{New1a7} & \textbf{New1a5}        & \textbf{New1c1}        & \textbf{New2a7}         \\ 
\hline
13                    &                 &                 &                 & New2d2          & New2a5          & New1a8          & New2d2          & \textbf{New1b1} & New1a8          & \textbf{New2d2} & \textbf{New1a5} & \textbf{New1b1} & \textbf{New2d2}        & \textbf{New2d3}        & \textbf{New1a5}         \\ 
\hline
14                    &                 &                 &                 & New1b1          & New1a4          & New2a3          & \textbf{New1b1} & New1b4          & New1c3          & \textbf{New1b1} & \textbf{New1b2} & \textbf{New1b2} & \textbf{New1b1}        & \textbf{New1a5}        & \textbf{New1a1}         \\ 
\hline
15                    &                 &                 &                 & New2a2          & New1c2          & \textbf{New2a1} & New2a2          & \textbf{New2a3} & \textbf{New2d1} & New2a2          & \textbf{New1b1} & \textbf{New2a1} & \textbf{New2a2}        & \textbf{New1b2}        & \textbf{New2a6}         \\ 
\hline
16                    &                 &                 &                 &                 &                 &                 & New1b3          & Age in months   & New2a4          & \textbf{New1b3} & New1b4          & \textbf{New1c1} & \textbf{New1b3}        & \textbf{New1b1}        & \textbf{New2d3}         \\ 
\hline
17                    &                 &                 &                 &                 &                 &                 & New2d3          & New2a5          & \textbf{New1a7} & \textbf{New2d3} & \textbf{New2a3} & \textbf{New2b}  & \textbf{New2d3}        & \textbf{New1b4}        & \textbf{New1a4}         \\ 
\hline
18                    &                 &                 &                 &                 &                 &                 & \textbf{New2a3} & New1a4          & Gender          & \textbf{New2a3} & Age in months   & \textbf{New2d2} & \textbf{New2a3}        & \textbf{New2a3}        & \textbf{New2c}          \\ 
\hline
19                    &                 &                 &                 &                 &                 &                 & New1b4          & New1c2          & \textbf{New2a1} & New1b4          & \textbf{New1b3} & \textbf{New2a4} & \textbf{New1b4}        & \textbf{Age in months} & \textbf{New2d4}         \\ 
\hline
20                    &                 &                 &                 &                 &                 &                 & New1c3          & Gender          & New1a6          & New1c3          & \textbf{New2a4} & New1a6          & \textbf{New1c3}        & \textbf{New1b3}        & \textbf{Age in months}  \\ 
\hline
21                    &                 &                 &                 &                 &                 &                 &                 &                 &                 & \textbf{New2a4} & New1a1          & \textbf{New2d4} & \textbf{New2a4}        & \textbf{New1c3}        & \textbf{New2a2}         \\ 
\hline
22                    &                 &                 &                 &                 &                 &                 &                 &                 &                 & \textbf{New1a4} & \textbf{New2a5} & \textbf{New2d1} & \textbf{New1a4}        & \textbf{New2a4}        & \textbf{New1b1}         \\ 
\hline
23                    &                 &                 &                 &                 &                 &                 &                 &                 &                 & Age in months   & \textbf{New1a4} & New1c2          & \textbf{Age in months} & \textbf{New1a1}        & \textbf{New2a4}         \\ 
\hline
24                    &                 &                 &                 &                 &                 &                 &                 &                 &                 & \textbf{New2a5} & New1c2          & \textbf{New2c}  & \textbf{New2a5}        & \textbf{New2a5}        & \textbf{New1b4}         \\ 
\hline
25                    &                 &                 &                 &                 &                 &                 &                 &                 &                 & New1a1          & Gender          & \textbf{New2d3} & \textbf{New1a1}        & \textbf{New1a4}        & \textbf{New1c1}         \\ 
\hline
26                    &                 &                 &                 &                 &                 &                 &                 &                 &                 &                 &                 &                 & \textbf{New1a8}        & \textbf{New1a8}        & \textbf{New1b2}         \\ 
\hline
27                    &                 &                 &                 &                 &                 &                 &                 &                 &                 &                 &                 &                 & \textbf{New1a2}        & \textbf{New1a2}        & \textbf{New2d2}         \\ 
\hline
28                    &                 &                 &                 &                 &                 &                 &                 &                 &                 &                 &                 &                 & \textbf{New1c2}        & \textbf{New1c2}        & \textbf{New2b}          \\ 
\hline
29                    &                 &                 &                 &                 &                 &                 &                 &                 &                 &                 &                 &                 & \textbf{New1a6}        & \textbf{New1a6}        & \textbf{New2d1}         \\ 
\hline
30                    &                 &                 &                 &                 &                 &                 &                 &                 &                 &                 &                 &                 & \textbf{Gender}        & \textbf{Gender}        & \textbf{Gender}         \\
\hline
\end{tabular}
}
\label{selectedfeature}
\end{table*}

In the subsequent phase of our analysis, we utilized a set of predefined values for k, precisely 10, 15, 20, 25, and 30. After majority voting on each K value, we got the set of 4, 8, 11, 20, and 30 features, respectively, as shown in Table \ref{selectedfeature}.

We made a significant discovery: the optimal performance of our classifiers was achieved with a set of 20 features, which we found for K = 25. This finding underscores the importance of our feature selection approach in enhancing predictive performance. 
\begin{figure*}[!ht]
    \centering
    \includegraphics[width=0.3\textwidth]{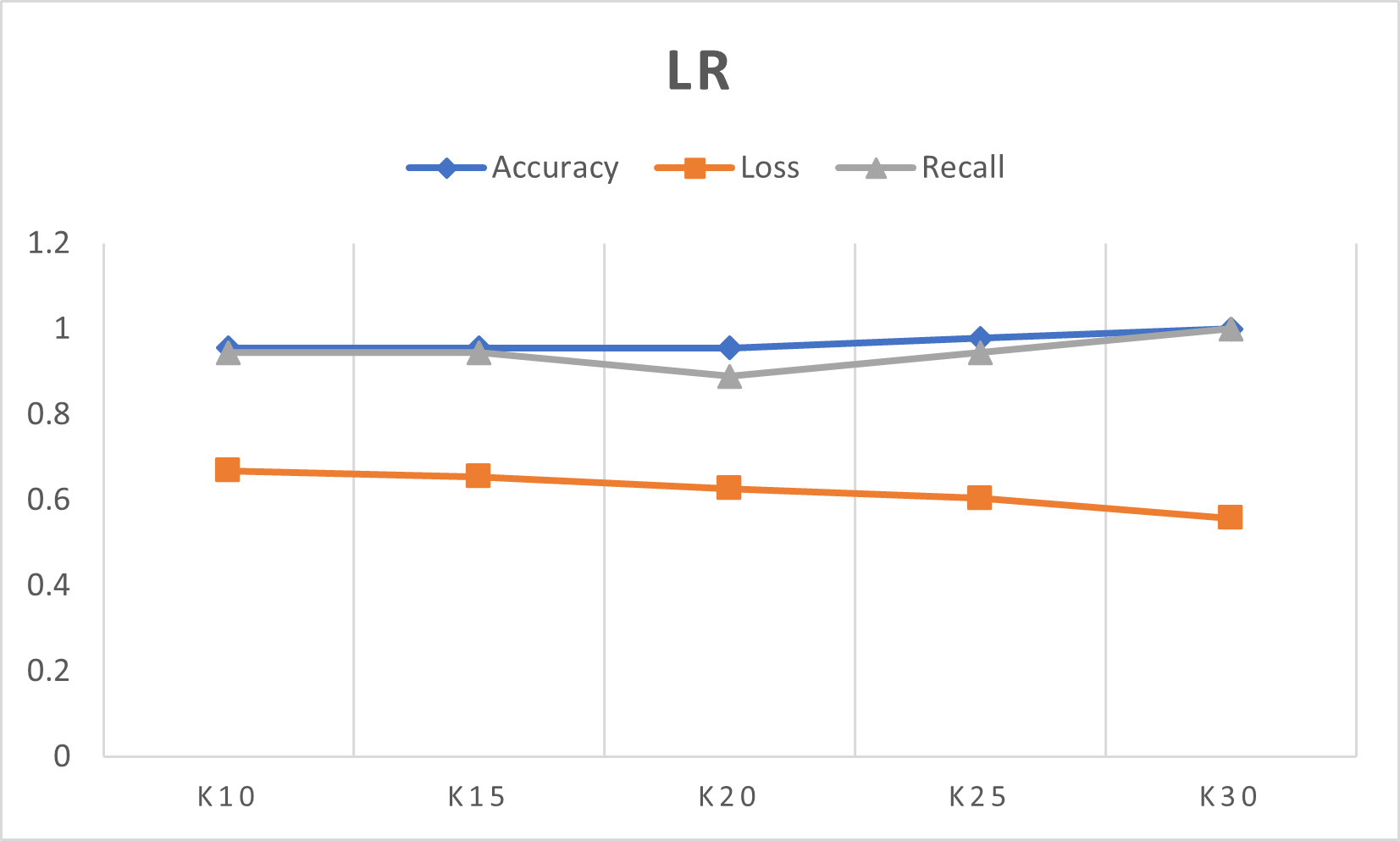}\hfill
    \includegraphics[width=0.3\textwidth]{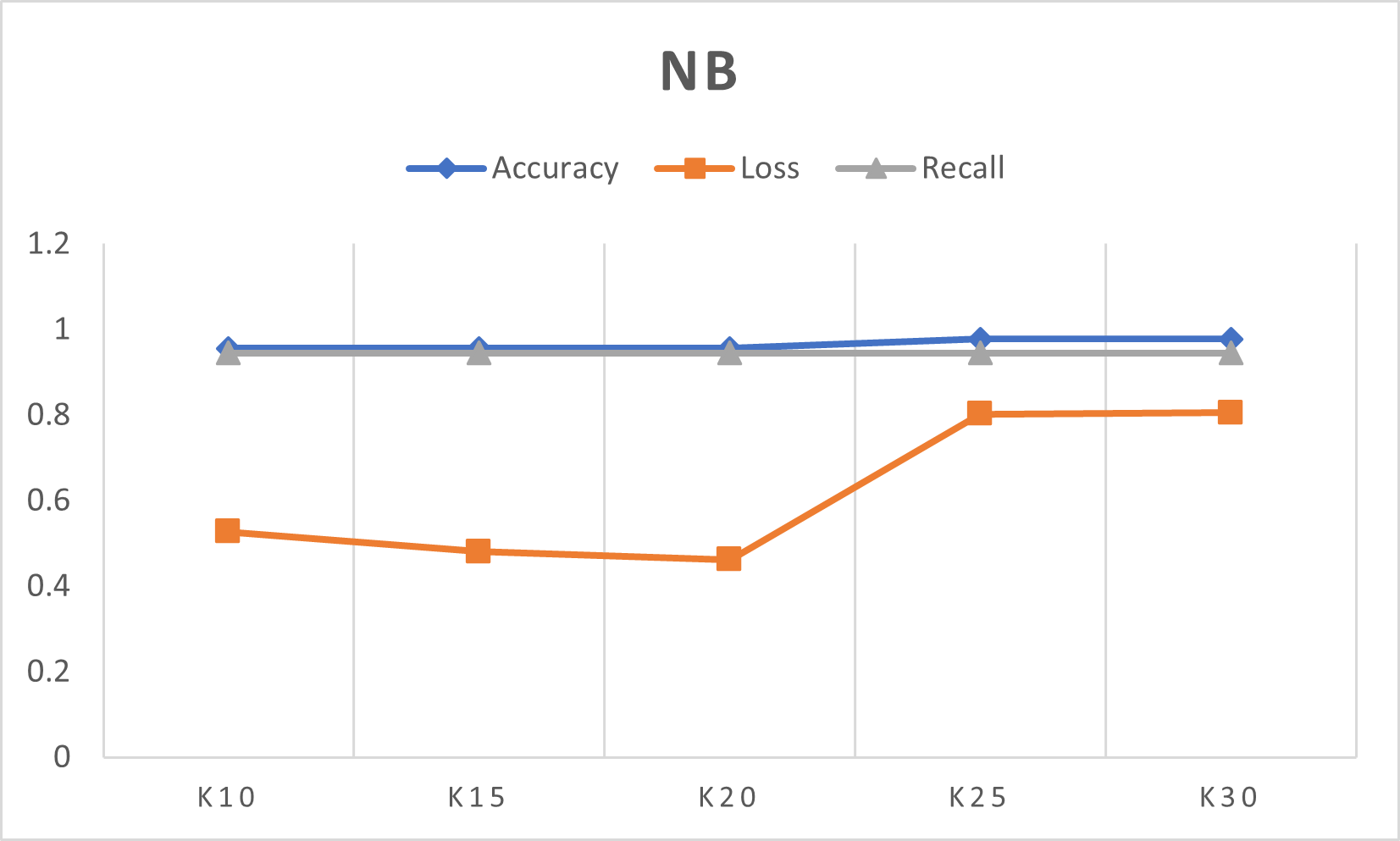}\hfill
    \includegraphics[width=0.3\textwidth]{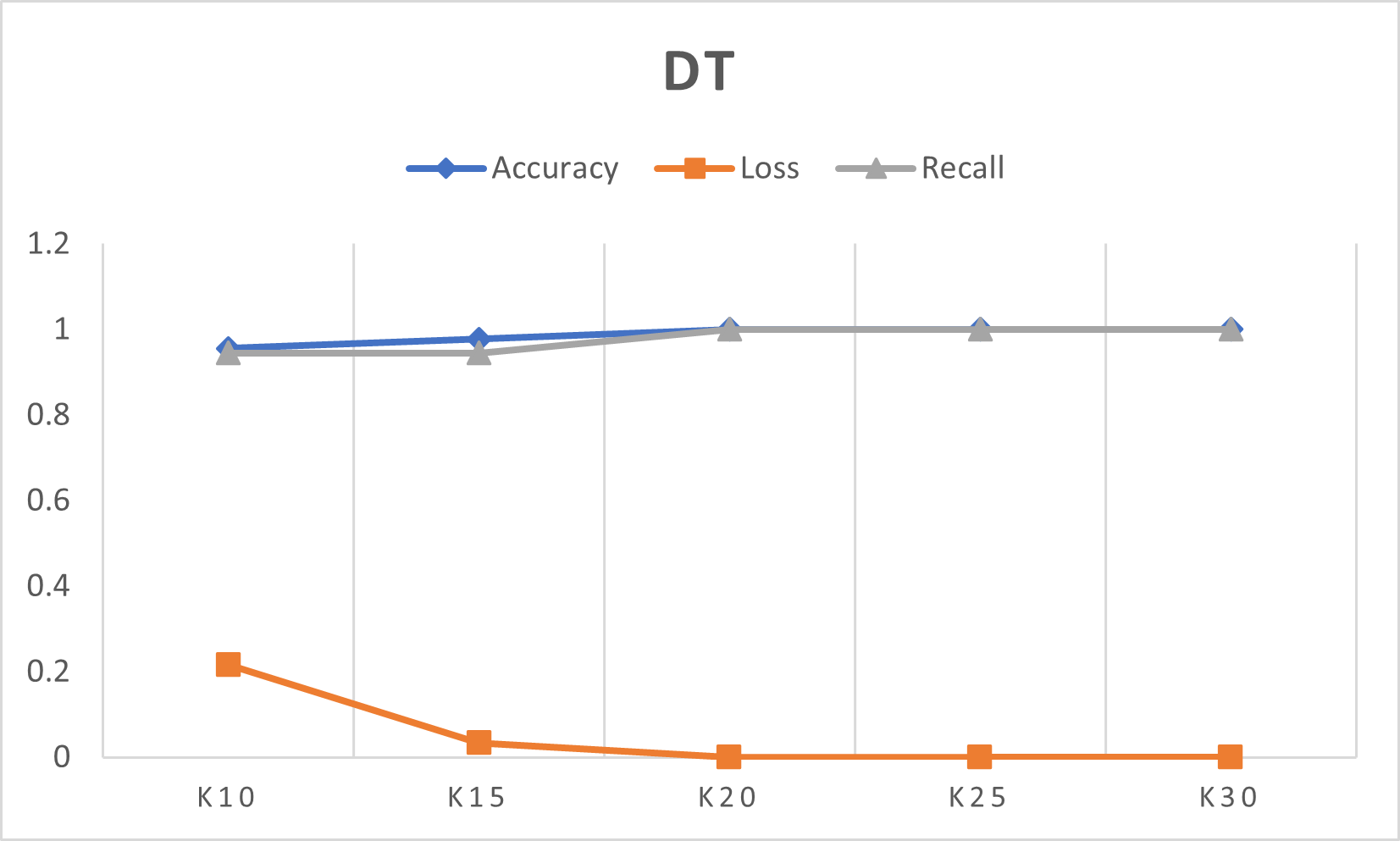}\\
    
    \includegraphics[width=0.3\textwidth]{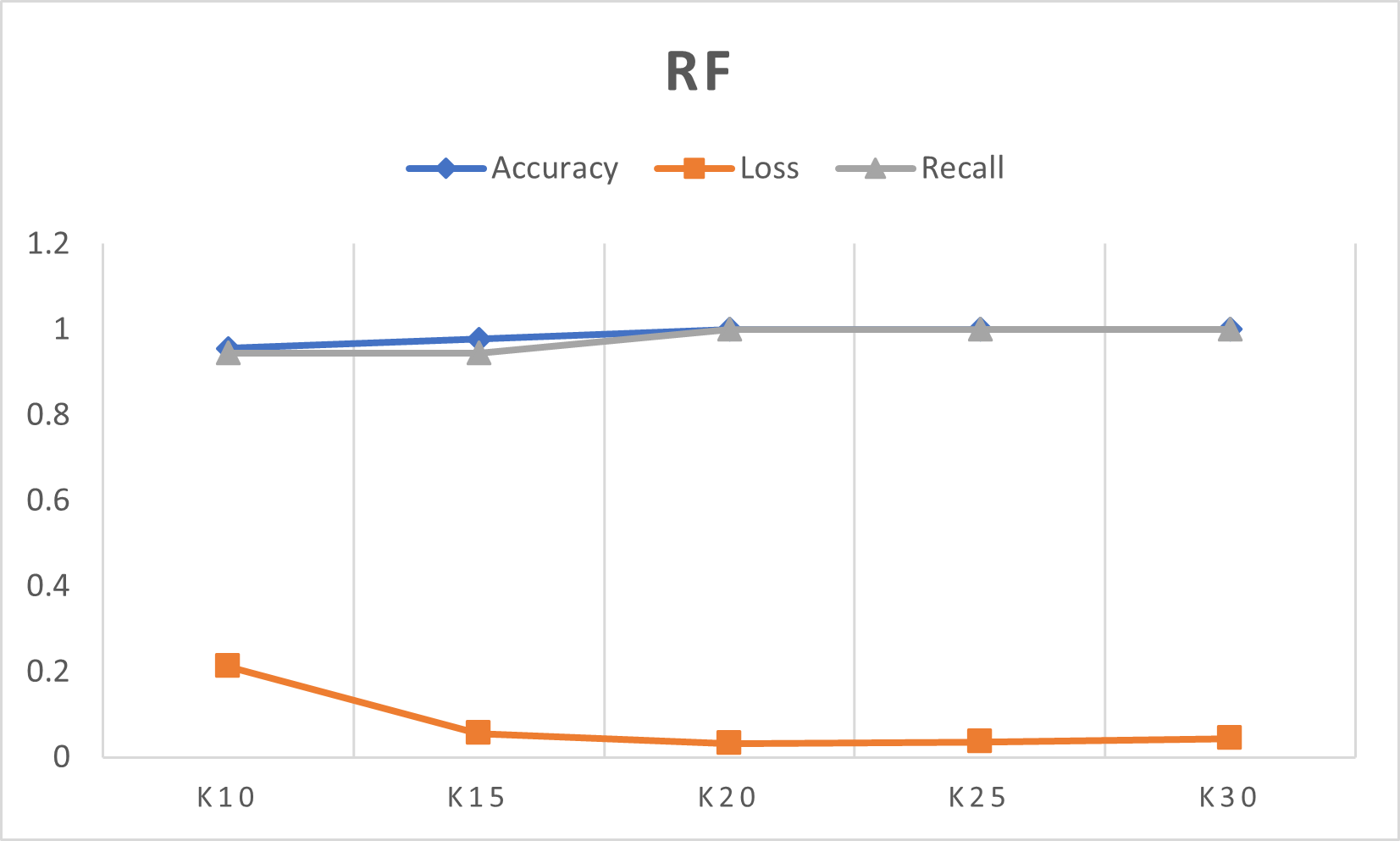}\hfill
    \includegraphics[width=0.3\textwidth]{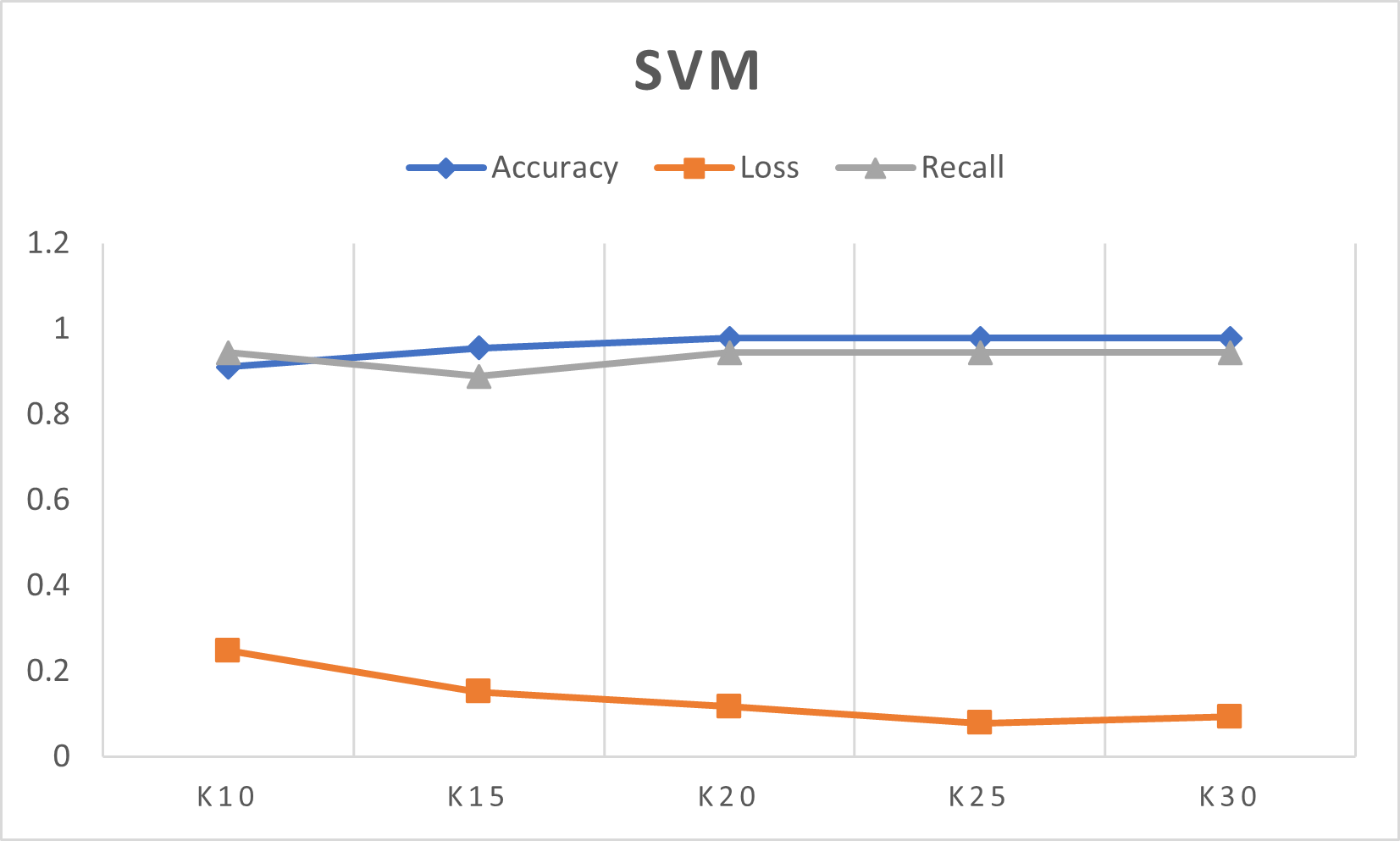}\hfill
    \includegraphics[width=0.3\textwidth]{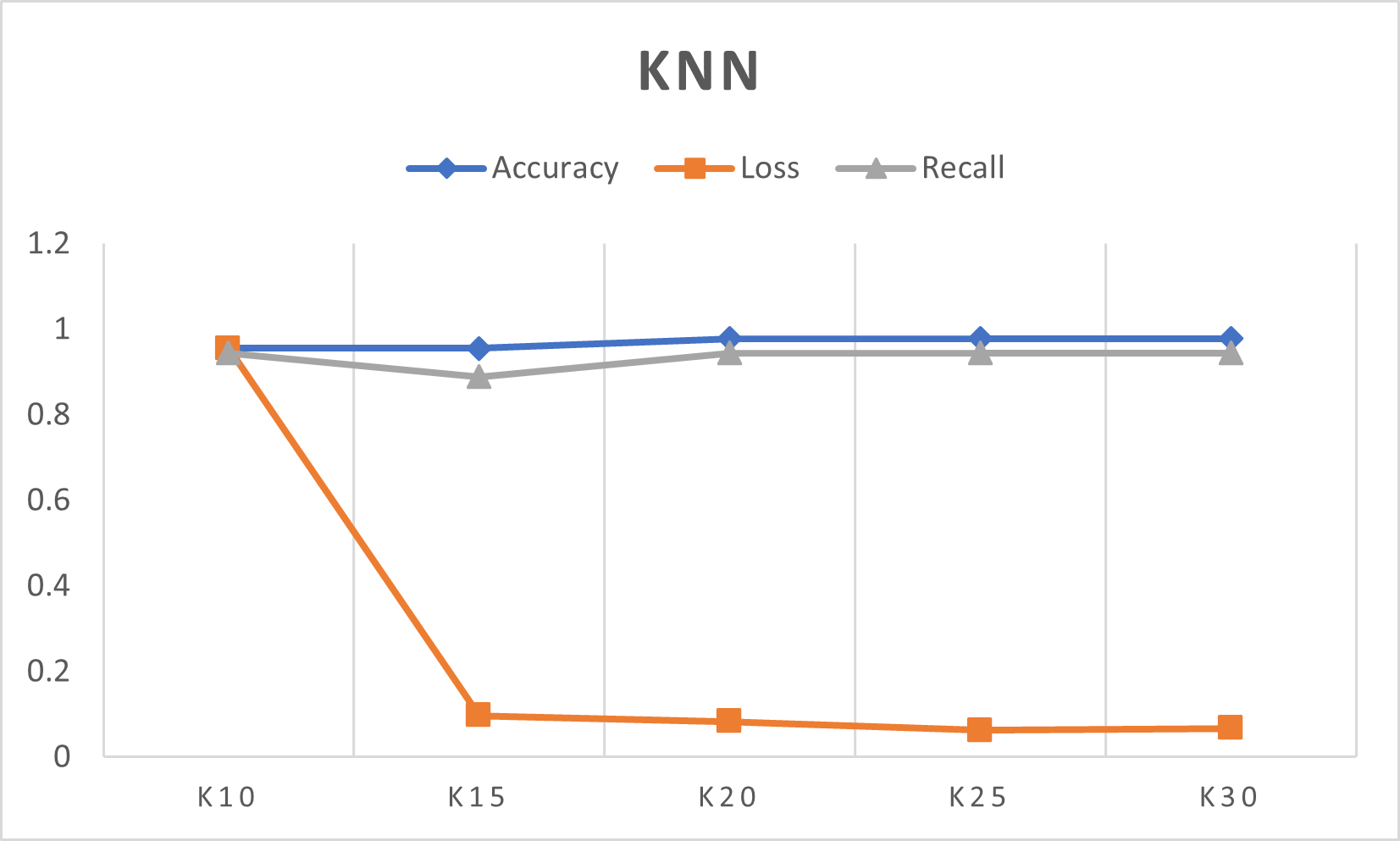}\\
    
    \includegraphics[width=0.3\textwidth]{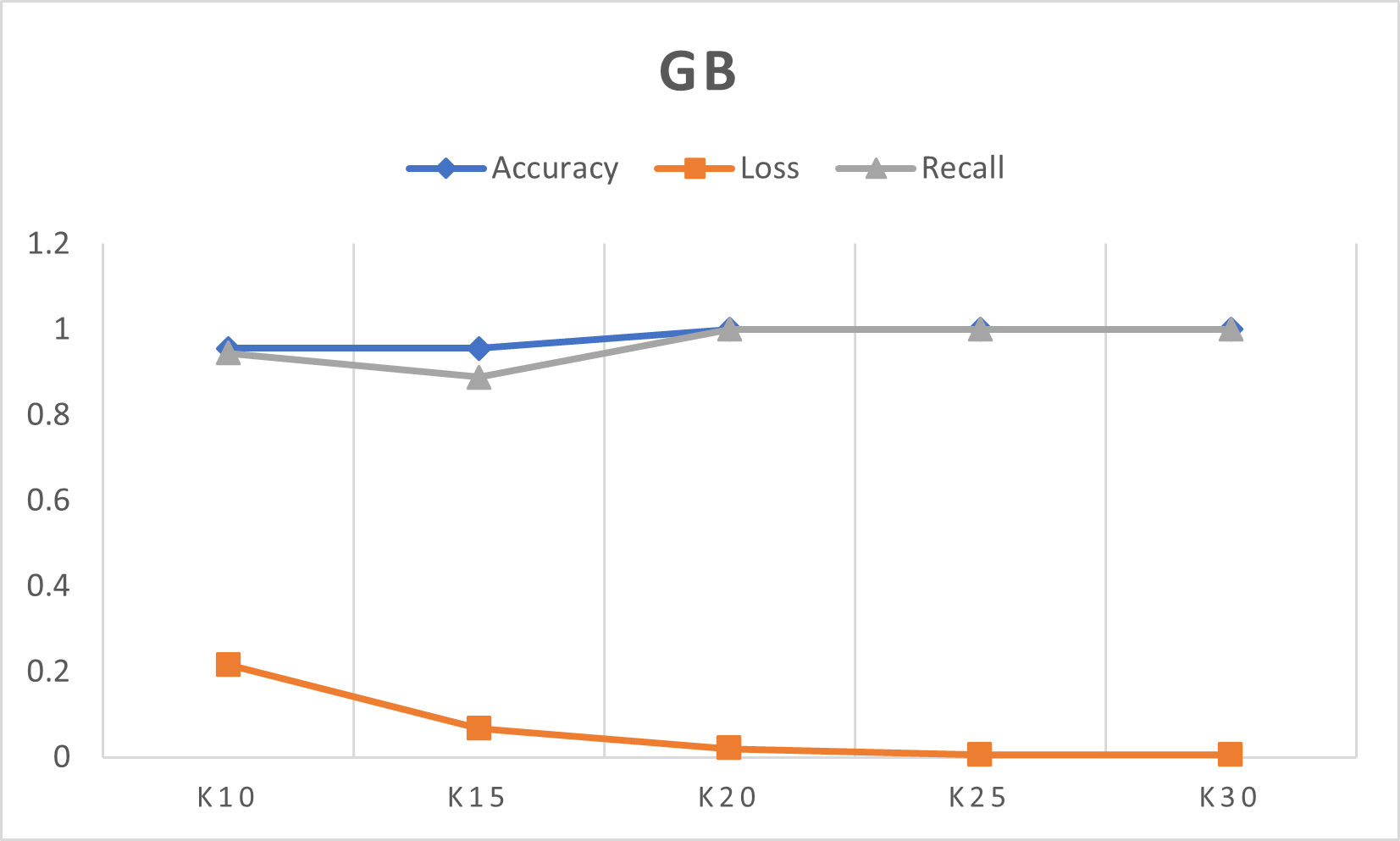}\hfill
    \includegraphics[width=0.3\textwidth]{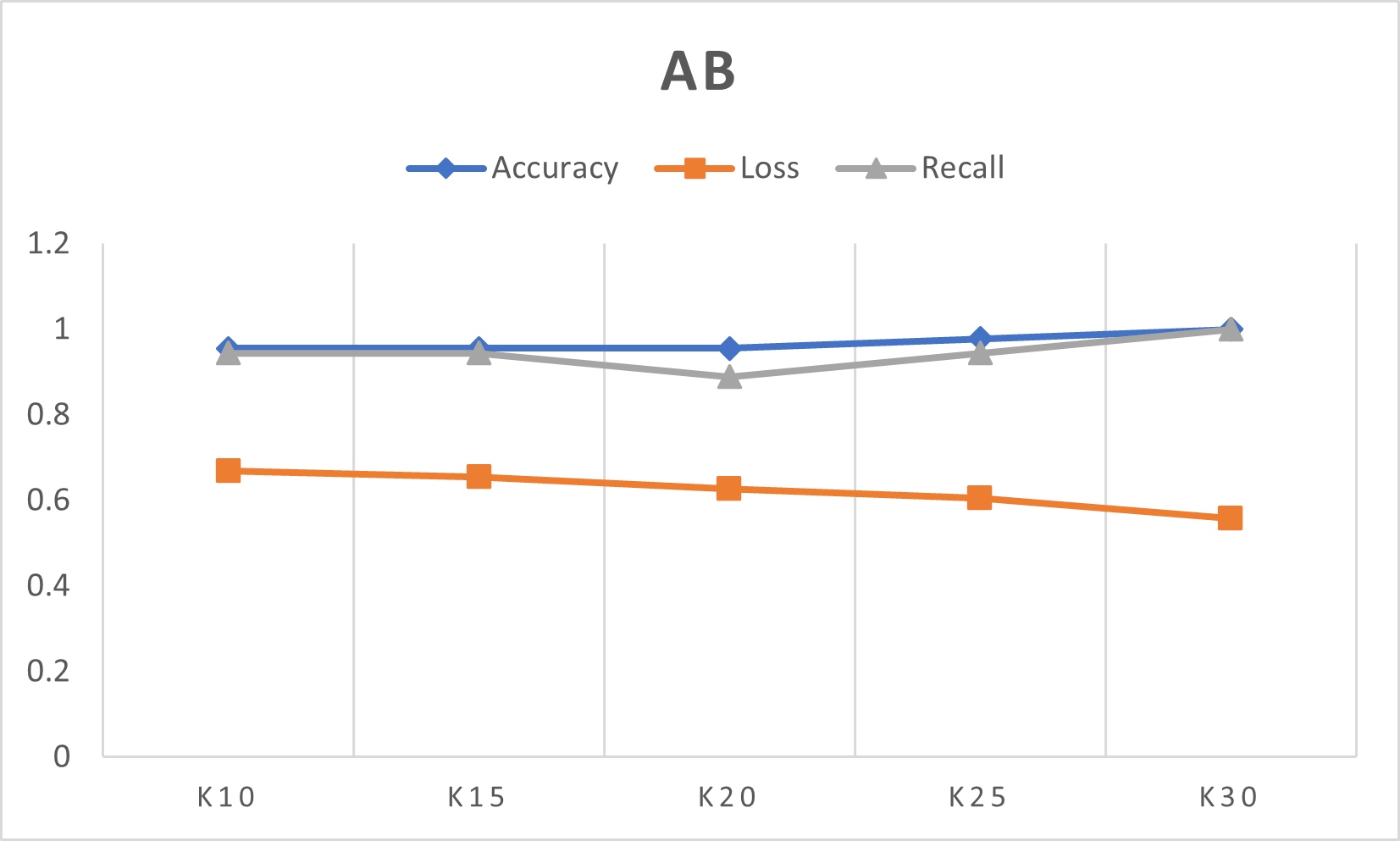}\hfill
    \includegraphics[width=0.3\textwidth]{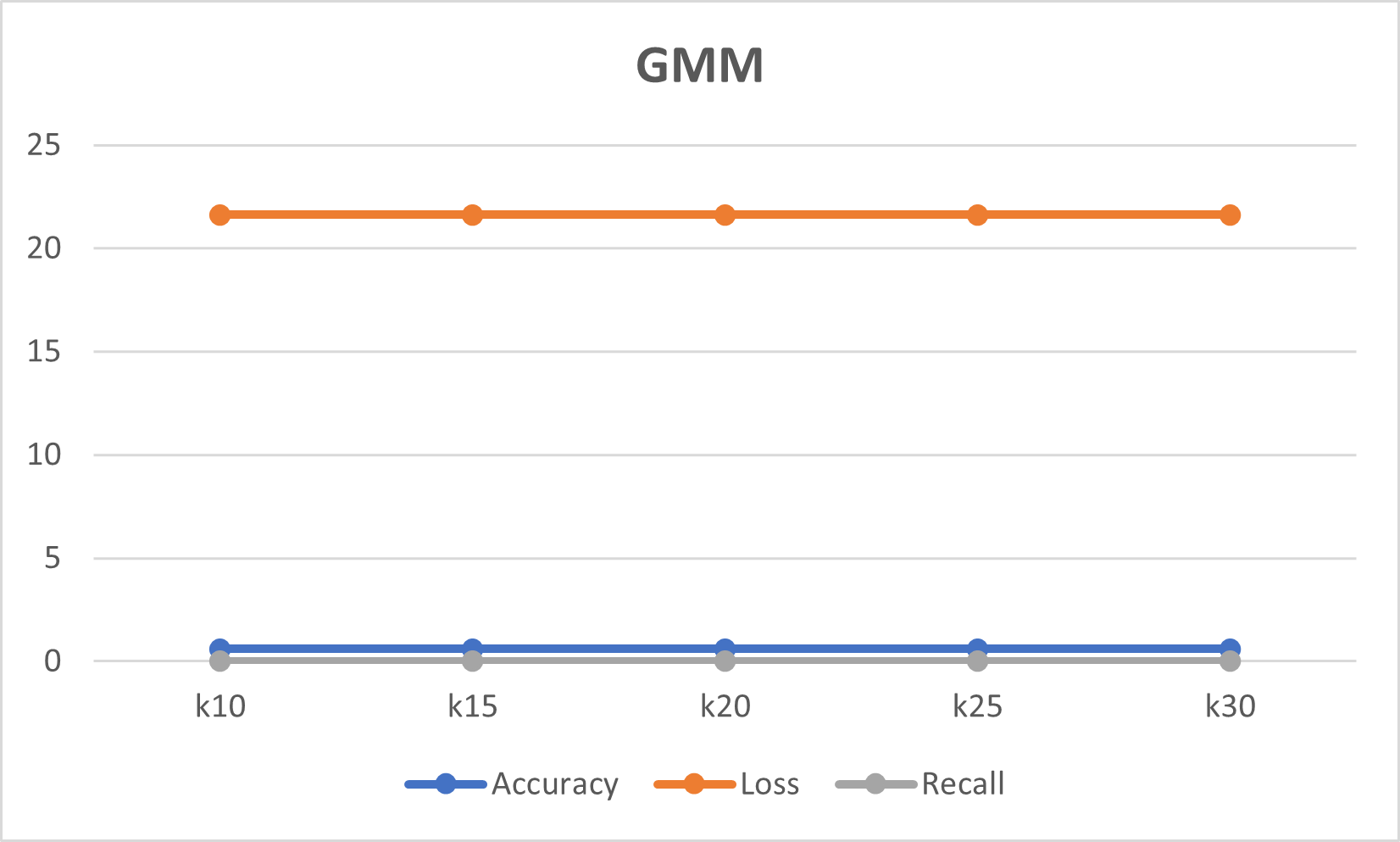}\\
    
    \includegraphics[width=0.3\textwidth]{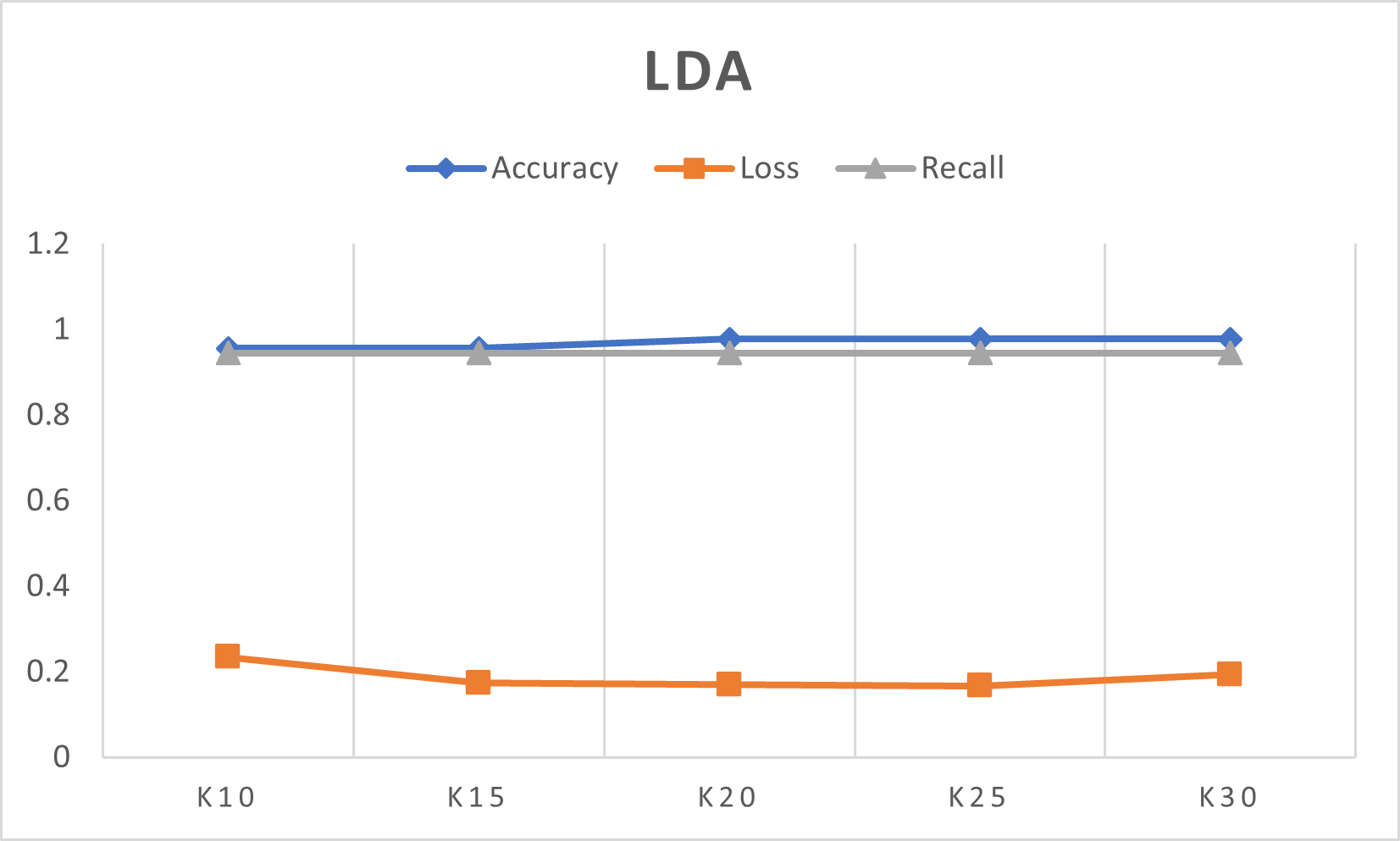}\hfill
    \includegraphics[width=0.3\textwidth]{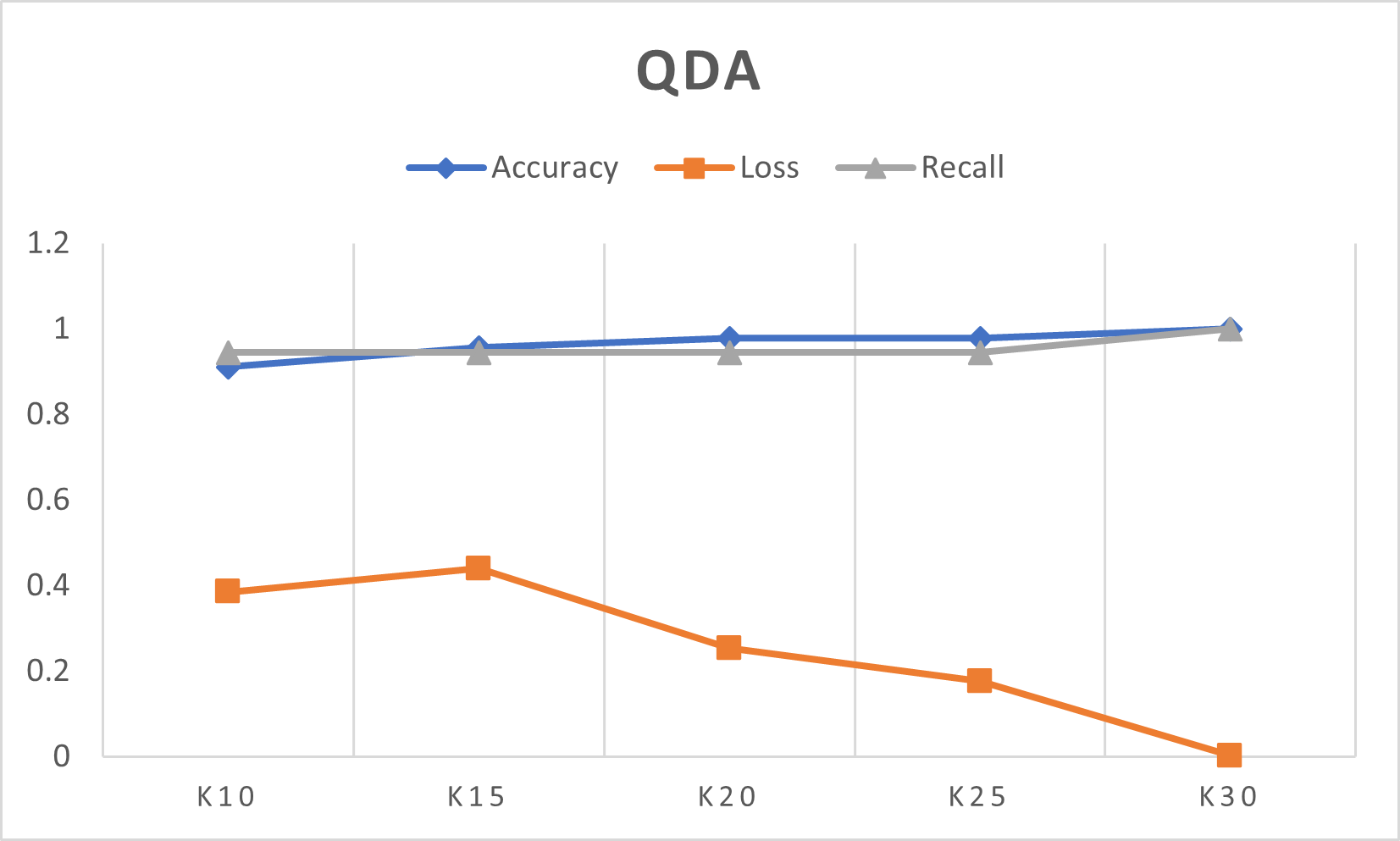}
    
    \caption{Performance Comparison of ML Models Across Various K Values}
    \label{fig:FS}
\end{figure*}
It became clear that increasing the maximum number of selected features up to K = 25 consistently improved classification accuracy, as shown in Figure \ref{fig:FS}. However, further increments after K = 25 in the number of features did not lead to additional improvements in accuracy, emphasizing the critical role of feature selection in our study.
\subsection{MACHINE LEARNING CLASSIFIERS} \label{PM}

In our study, various classifiers were trained on pre-processed data, including LR, NB, DT, RF, SVM, KNN, GB, AB, GMM, LDA, and QDA. Here, we discuss their mathematical underpinnings with some brief details:

\textbf{LR} \cite{menard2002applied} is a fundamental statistical method used to predict the likelihood of a dichotomous result in classification problems. It extends linear regression to classification scenarios by estimating the probability of a binary response, utilizing one or more independent predictors.

\[
P(Y=1) = \frac{1}{1 + e^{-(\beta_0 + \beta_1 X)}}
\]

\textbf{NB} \cite{bayes1968naive} is from a set of straightforward probabilistic algorithms that apply Bayes' theorem while presuming a high degree of independence among predictors. It's particularly suited for high-dimensional datasets.


\textbf{DT} \cite{song2015decision} are non-parametric models that apply supervised learning to perform classification and regression tasks. They construct the decision-making process as a tree-like model, segmenting the dataset based on the attributes.

\textbf{RF} \cite{biau2016random} is a composite method used for classification and regression, where it builds numerous DTs during training. The most frequent output from all the trees is taken as the final result for classification.

\[
E \leq \frac{\bar{\rho}(1 - \bar{s}^2)}{\bar{s}^2}
\]

\textbf{SVM} \cite{suthaharan2016support} encompass a group of supervised learning techniques suitable for classification, regression, and outlier detection tasks. These methods establish a hyperplane or multiple hyperplanes in a multidimensional space, facilitating tasks ranging from classification to regression and even identifying outliers. This method identifies the optimal hyperplane to distinguish between various classes, focusing on increasing the space between the nearest members of differing classes.

\[
\min_{\mathbf{w},b} \frac{1}{2}\mathbf{w}^\top\mathbf{w} + C\sum_{i=1}^n \xi_i
\]

\textbf{k-NN} \cite{peterson2009k} embodies a strategy of learning on-demand, also known as lazy learning, where predictions are generated by leveraging the proximity of data points in the feature space, with computational processes deferred until classification needs to be performed. It's one of the most straightforward ML algorithms, utilizing different distance measures such as Euclidean or Manhattan to identify the nearest neighbors.




\textbf{GB} \cite{natekin2013gradient} is a method in machine learning focused on addressing both regression and classification challenges. Predictive framework is created by aggregating several basic predictive models, with DT frequently employed, into a robust ensemble.

\[
F_m(x) = F_{m-1}(x) + \rho_m h_m(x)
\]

\textbf{AB} \cite{hastie2009multi}, is a machine learning technique that starts by applying a classifier to the original dataset. Subsequently, it trains further classifier instances, modifying the weights of incorrectly classified examples. This strategy guarantees that subsequent classifiers focus more on the more challenging examples to classify accurately.

\[
\alpha_t = \frac{1}{2} \ln \left(\frac{1 - \text{error}_t}{\text{error}_t}\right)
\]

\textbf{GMM} \cite{reynolds2009gaussian} is a statistical model that encapsulates the presence of subpopulations within an overall population, assuming that each subpopulation is normally distributed. GMMs can be used for finding clusters in the same manner as k-means. The likelihood of \textit{X} is given by

\[
P(X|\lambda) = \sum_{i=1}^k \pi_i \mathcal{N}(X|\mu_i,\Sigma_i)
\]

\textbf{LDA} \cite{xanthopoulos2013linear} is a technique employed in statistical analysis, pattern recognition, and ML that aims to identify a linear combination of features distinguishing or categorizing multiple classes. LDA makes the assumption that the probability density functions of the predictor variables are normally distributed. LDA calculates the discriminant scores with:

\[
Y = \mathbf{w}^\top \mathbf{X}
\]

\textbf{QDA} \cite{srivastava2007bayesian} is related to LDA. Still, it allows for the possibility that the variance-covariance matrix is not identical across the groups, providing a more flexible classifier.

\subsubsection{ML Model Optimization}
\mbox{}\\
\begin{table*}[!htbp]
\centering
\caption{Hyperparameter Descriptions}
\resizebox{\columnwidth * 2}{!}
{
\begin{tabular}{|p{0.1\textwidth}|p{0.43\textwidth}|p{0.27\textwidth}|p{0.2\textwidth}|}
\hline
\textbf{Classifiers} & \textbf{Possible Hyperparameters} & \textbf{Best Hyperparameters} & \textbf{Best average score on all CV folds} \\ \hline \hline

LR & 
\textbf{penalty}: ['l1'],\newline
\textbf{C}: [0.01, 0.1, 1.0, 10.0],\newline
\textbf{dual}: [False], \newline
\textbf{tol}: [0.0001, 0.001, 0.01],\newline 
\textbf{fit\_intercept}: [True, False],\newline 
\textbf{intercept\_scaling}: [1, 2, 5],\newline 
\textbf{class\_weight}: [None, 'balanced'],\newline
\textbf{solver}: ['saga'],\newline 
\textbf{max\_iter}: [1000, 2000, 5000],\newline 
\textbf{l1\_ratio}: [None] &

\textbf{penalty}: l1,\newline
\textbf{C}: 0.1,\newline
\textbf{dual}: False,\newline
\textbf{tol}: 0.01,\newline
\textbf{fit\_intercept}: True,\newline
\textbf{intercept\_scaling}: 1,\newline
\textbf{class\_weight}: balanced,\newline
\textbf{solver}: saga,\newline
\textbf{max\_iter}: 5000,\newline
\textbf{l1\_ratio}: None &

0.8978\\ \hline

NB & NO tunable parameter for naive bayes & NA & NA \\ \hline

DT &
\textbf{criterion}: ['gini', 'entropy'],\newline
\textbf{max\_depth}: [None, 5, 10, 20],\newline
\textbf{min\_samples\_split}: [2, 5, 10],\newline
\textbf{min\_samples\_leaf}: [1, 2, 4],\newline
\textbf{max\_features}: [None, 'sqrt', 'log2'] &

\textbf{criterion}: gini,\newline
\textbf{max\_depth}: 5,\newline
\textbf{min\_samples\_split}: 2,\newline
\textbf{min\_samples\_leaf}: 1, \newline
\textbf{max\_features}: None &

0.9289\\ \hline

RF &
\textbf{n\_estimators}: [100, 200, 300],\newline
\textbf{criterion}: ['gini', 'entropy'],\newline
\textbf{max\_depth}: [None, 5, 10],\newline
\textbf{min\_samples\_split}: [2, 5, 10],\newline
\textbf{min\_samples\_leaf}: [1, 2, 4],\newline
\textbf{max\_features}: ['sqrt', 'log2'],\newline
\textbf{ccp\_alpha}: [0.0, 0.1, 0.2] &

\textbf{n\_estimators}: 200,\newline
\textbf{criterion}: gini,\newline
\textbf{max\_depth}: 10,\newline
\textbf{min\_samples\_split}: 5,\newline
\textbf{min\_samples\_leaf}: 4,\newline
\textbf{max\_features}: log2,\newline
\textbf{ccp\_alpha}: 0.0 &

0.9422\\ \hline

SVM &
\textbf{C}: [0.1, 1, 10],\newline
\textbf{kernel}: ['linear', 'rbf', 'sigmoid'],\newline
\textbf{gamma}: ['scale', 'auto'] &

\textbf{C}: 10, \newline
\textbf{kernel}: 'rbf',\newline
\textbf{gamma}: 'scale' &

0.9022\\ \hline

KNN &
\textbf{n\_neighbors}: [3, 5, 7],\newline
\textbf{weights}: ['uniform', 'distance'],\newline
\textbf{algorithm}: ['auto', 'ball\_tree', 'kd\_tree', 'brute'],\newline
\textbf{p}: [1, 2] &

\textbf{n\_neighbors}: 7,\newline
\textbf{weights}: 'distance'\newline
\textbf{algorithm}: 'ball\_tree', \newline
\textbf{p}: 1 &

0.9022\\ \hline

GB &
\textbf{learning\_rate}: [0.1, 0.01, 0.001],\newline
\textbf{n\_estimators}: [100, 200, 300],\newline
\textbf{max\_depth}: [3, 4, 5],\newline
\textbf{subsample}: [0.8, 1.0],\newline
\textbf{min\_samples\_split}: [2, 4, 6],\newline
\textbf{min\_samples\_leaf}: [1, 2, 3] &

\textbf{learning\_rate}: 0.01,\newline
\textbf{n\_estimators}: 100,\newline
\textbf{max\_depth}: 3,\newline
\textbf{subsample}: 0.8,\newline
\textbf{min\_samples\_split}: 4,\newline
\textbf{min\_samples\_leaf}: 3 &

0.9378\\ \hline

AB &
\textbf{n\_estimators}: [50, 100, 200],\newline
\textbf{learning\_rate}: [0.01, 0.1, 1.0],\newline
\textbf{algorithm}: ['SAMME', 'SAMME.R'] & 

\textbf{n\_estimators}: 50,\newline
\textbf{learning\_rate}: 1.0,\newline
\textbf{algorithm}: 'SAMME' & 

0.9155\\ \hline

GMM &
\textbf{n\_components}: [2, 3, 4, 5],\newline
\textbf{covariance\_type}: ['spherical', 'tied', 'diag', 'full']  &

\textbf{n\_components}: 2,\newline
\textbf{covariance\_type}: 'diag'  &

0.9156\\ \hline

LDA &
\textbf{solver}: ['lsqr', 'eigen'],\newline
\textbf{shrinkage}: ['auto', None, 0.5, 1.0],\newline
\textbf{n\_components}: [None, 1] &

\textbf{solver}: 'lsqr',\newline
\textbf{shrinkage}: None,\newline
\textbf{n\_components}: None &

0.9067\\ \hline

QDA &
\textbf{reg\_param}: [0.0, 0.1, 0.2, 0.3, 0.4, 0.5, 0.6, 0.7, 0.8, 0.9, 1.0] &

\textbf{reg\_param}: 0.1 &

0.8933\\ \hline

\end{tabular}
}
\label{Table2}
\end{table*}

Following the initial evaluation of our classifiers using 5-fold cross-validation, we embarked on a parameter tuning phase to refine the models further and enhance their predictive performance. To automate this crucial process, we employed the GridSearchCV method from the sklearn library, a tool known for its precision and effectiveness. The tool thoroughly assesses all potential combinations of parameter values for a given estimator by performing an extensive search within a defined range. It uses cross-validation to appraise the efficacy of each parameter set.

The grid search was executed over the classifiers, with the parameter grid defined in Table \ref{Table2}. The search used 5-fold cross-validation to ensure a comprehensive evaluation, aligning with our initial model assessment method. This approach facilitated a detailed comparison of model performance across a spectrum of hyperparameter combinations. The search yielded many results, encapsulating the performance metrics for each parameter set. It offered a quantifiable measure of how well the model, configured with the optimal hyperparameters found by GridSearchCV, is expected to perform on unseen data based on the cross-validation procedure. This score was a significant indicator of the model's generalization ability, assuming the data represents the problem space.

The exhaustive search identified optimal parameters that maximized our classifier's performance. These parameters were extracted from the grid search results, offering insights into the most effective configuration for our predictive model. Utilizing this set, we retrain the classifiers, employing the entire dataset to finalize and prepare the model.

\section{WEB-BASED APPLICATION FOR ASD} \label{WBA}
\begin{figure*}[!htbp]
    \centering
    \begin{subfigure}[b]{0.49\linewidth}
    \centering
    \includegraphics[width=\linewidth, height=0.2\textheight, keepaspectratio]{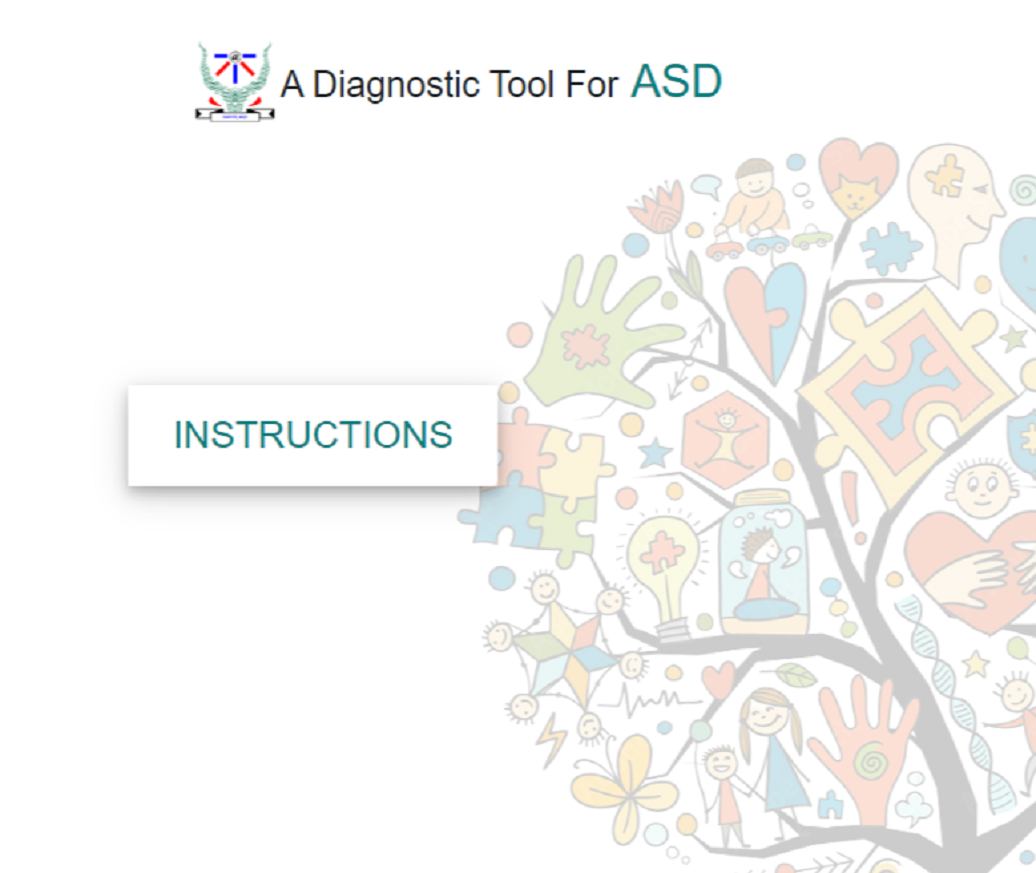}
    \end{subfigure}
    \centering
    \hfill 
    \begin{subfigure}[b]{0.49\linewidth}
    \centering
        \includegraphics[width=\linewidth, height=0.2\textheight, keepaspectratio]{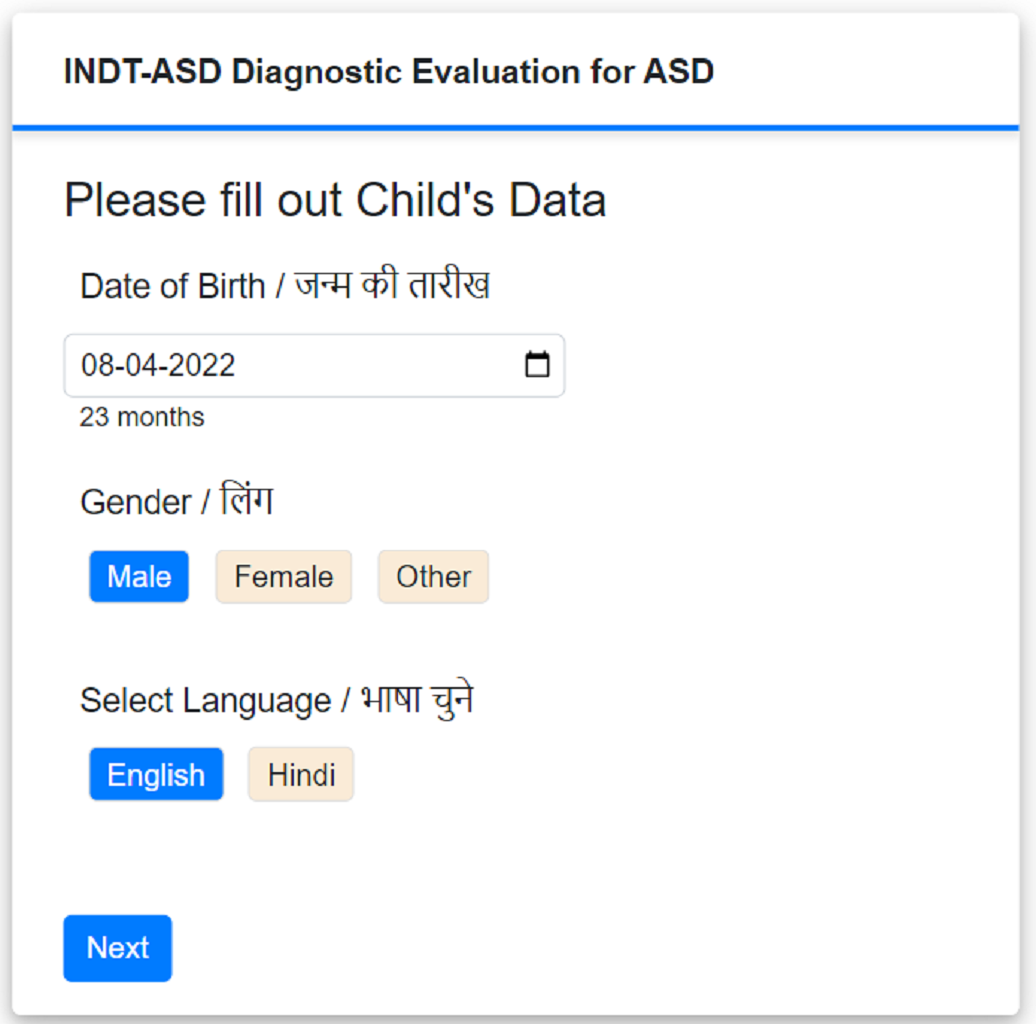}
    \end{subfigure}
    
    \begin{subfigure}[b]{0.49\linewidth}
    \centering
    \includegraphics[width=\linewidth, height=0.5\textheight, keepaspectratio]{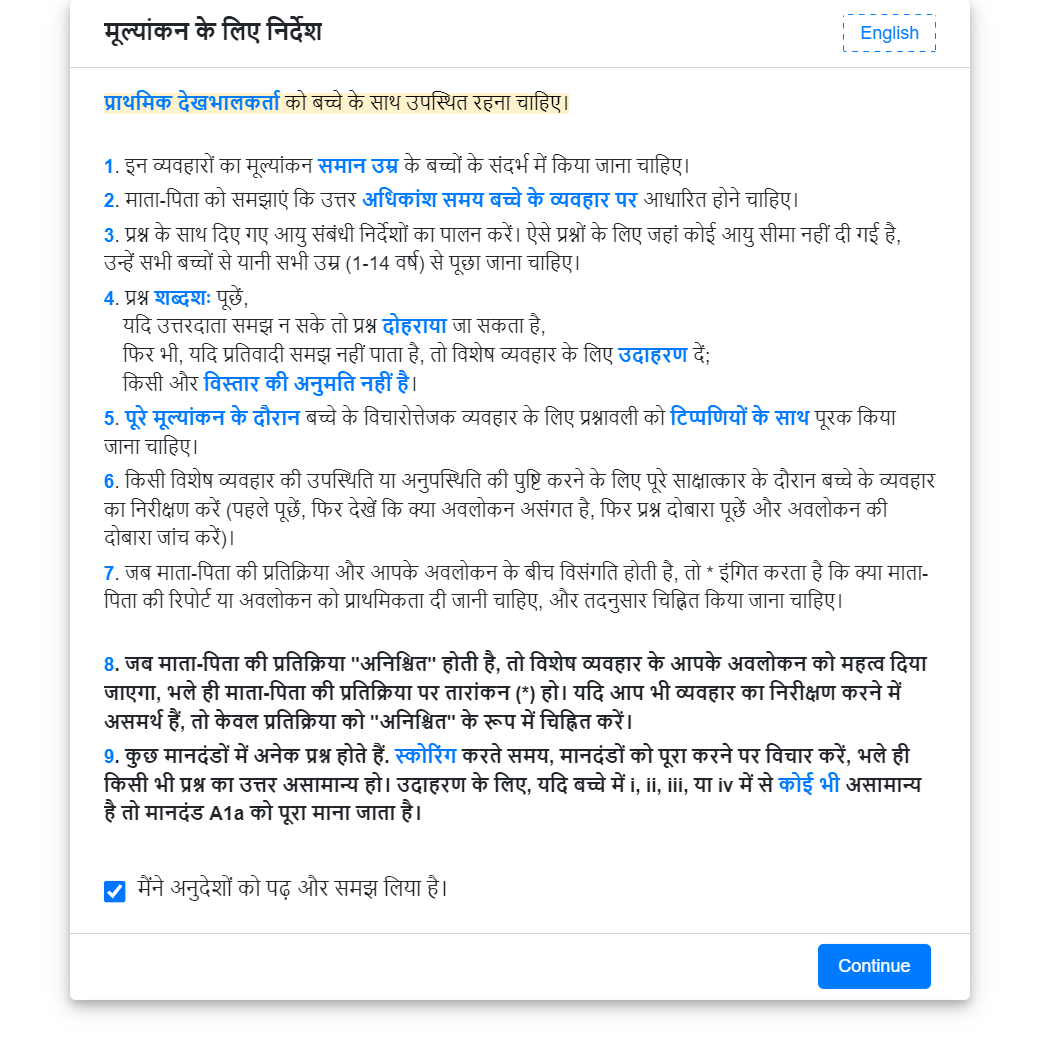}
    \end{subfigure}
    \hfill
    \begin{subfigure}[b]{0.49\linewidth}
    \centering
    \includegraphics[width=\linewidth, height=0.5\textheight, keepaspectratio]{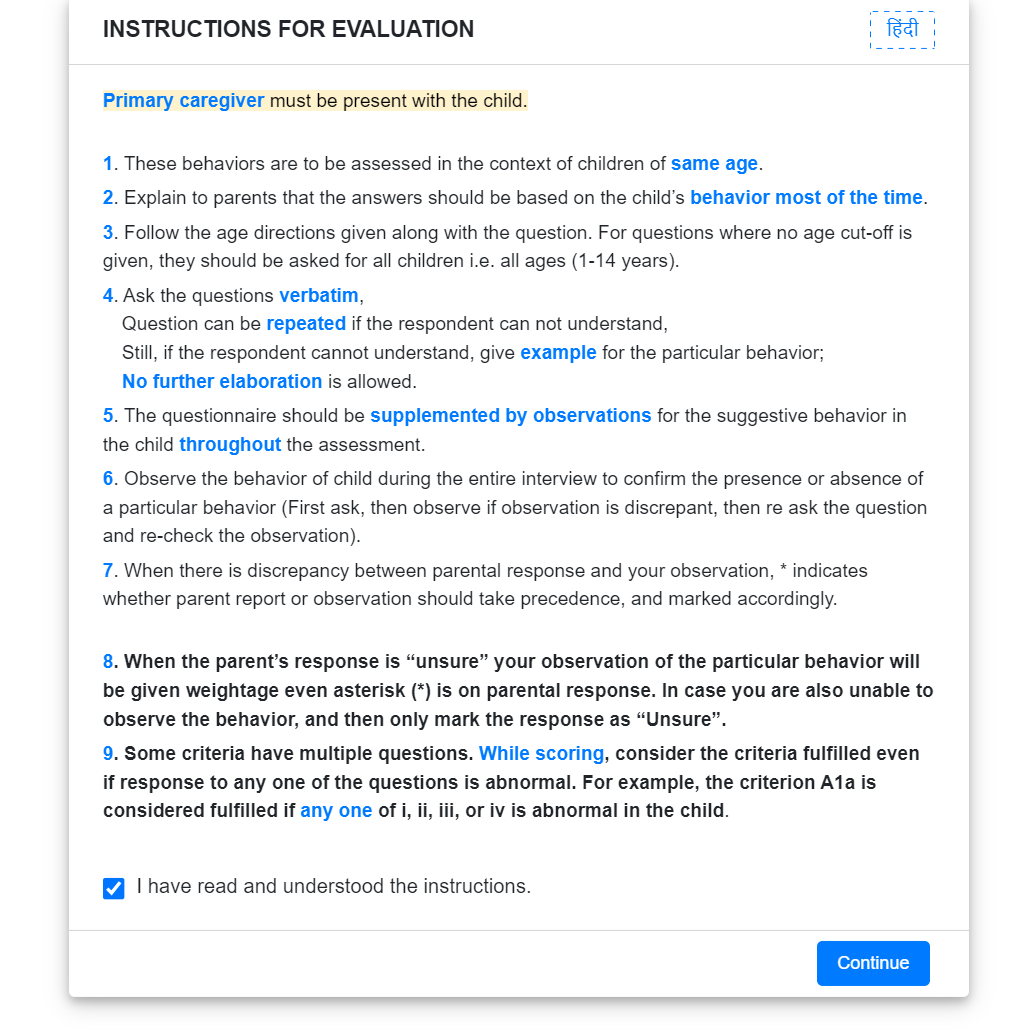}
    \end{subfigure}

    \begin{subfigure}[b]{0.49\linewidth}
    \centering
    \includegraphics[width=\linewidth, height=0.45\textheight, keepaspectratio]{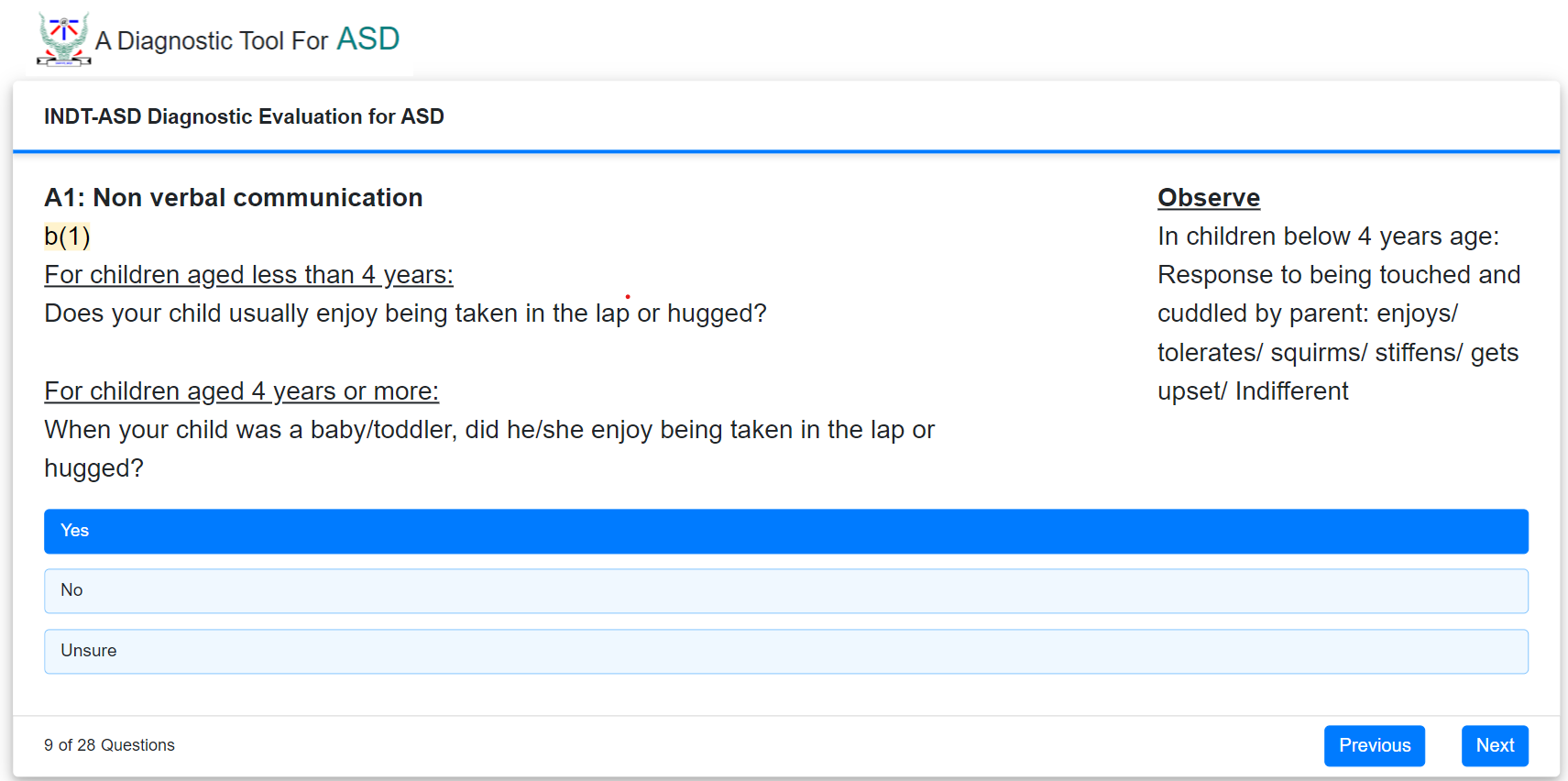}
    \end{subfigure}
    \hfill
    \begin{subfigure}[b]{0.49\linewidth}
    \centering
    \includegraphics[width=\linewidth, height=0.45\textheight, keepaspectratio]{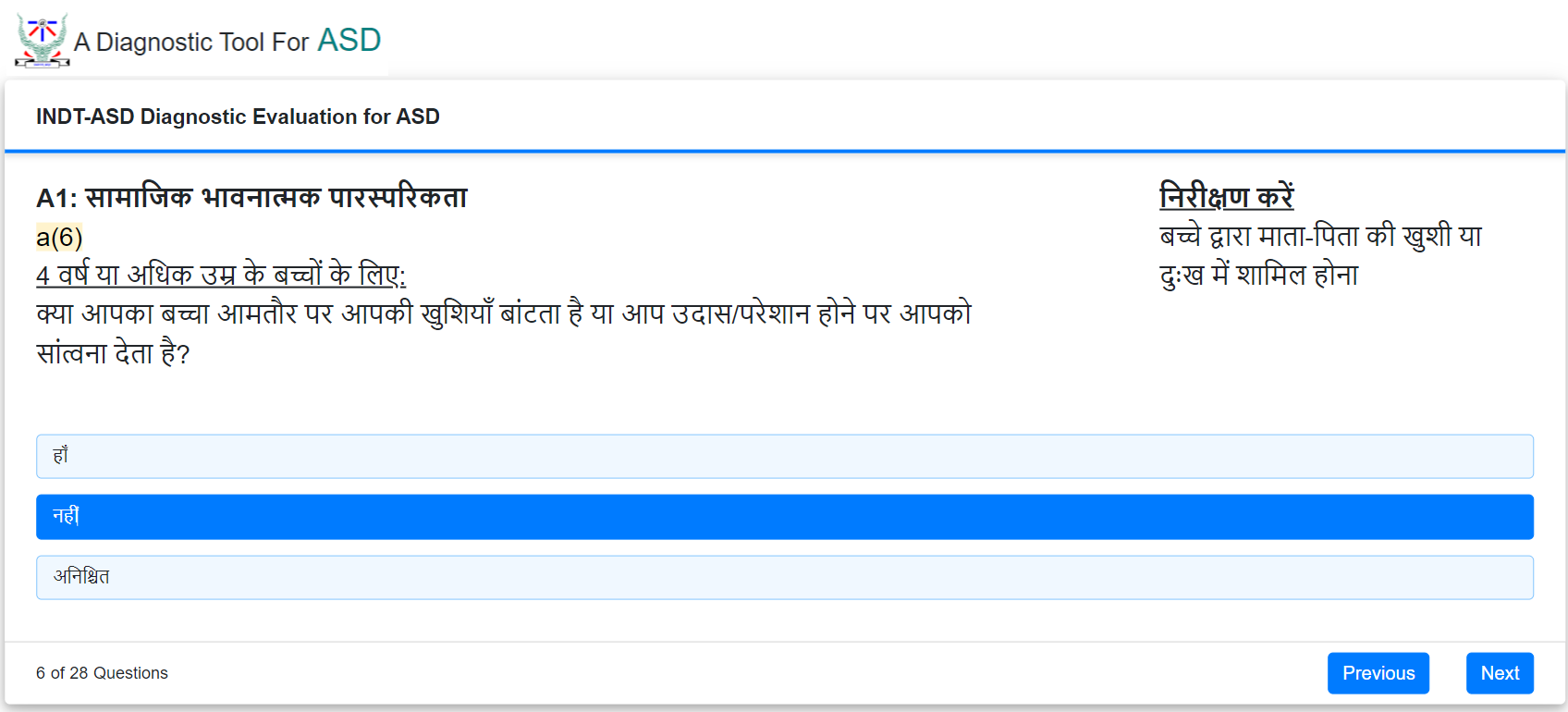}
    \end{subfigure}
    \hfill
    \begin{subfigure}[b]{0.49\linewidth}
    \centering
    \includegraphics[width=\linewidth, height=0.25\textheight, keepaspectratio]{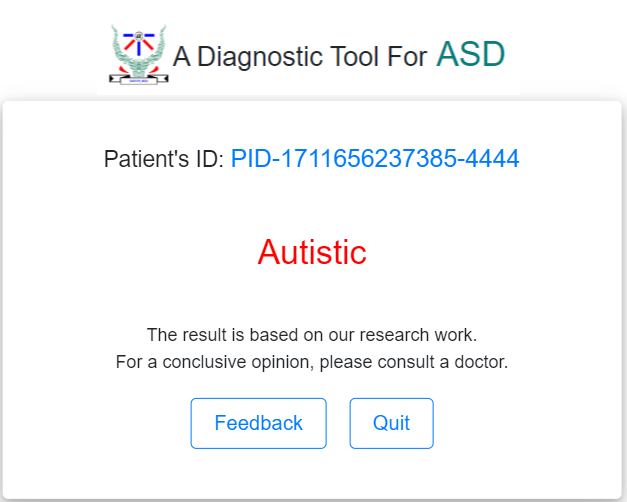}
    \end{subfigure}
    \caption{Web-Based Application Screenshots}
    \label{web-app}
\end{figure*}

We have developed an application aimed at early detection of autism. This tool uses the most effective model identified from our research, ensuring high reliability. It utilizes all the questions from the AIIMS-modified INCLEN (AMI) questionnaire. It is designed for parents, guardians, or doctors of an autistic patient; the application is very user-friendly. Recognizing India's linguistic diversity, it is available in both Hindi and English, catering to a broad audience. This strategic choice addresses many linguistic barriers, thus expanding its reach and utility. Our application serves as a valuable resource for early detection, offering families a pathway to seeking timely intervention. Figure \ref{web-app} displays the ASD tool.

\section{RESULT} \label{R}

\begin{figure*}
    \centering
    \includegraphics[width=1\linewidth, height=0.35\textheight,keepaspectratio]{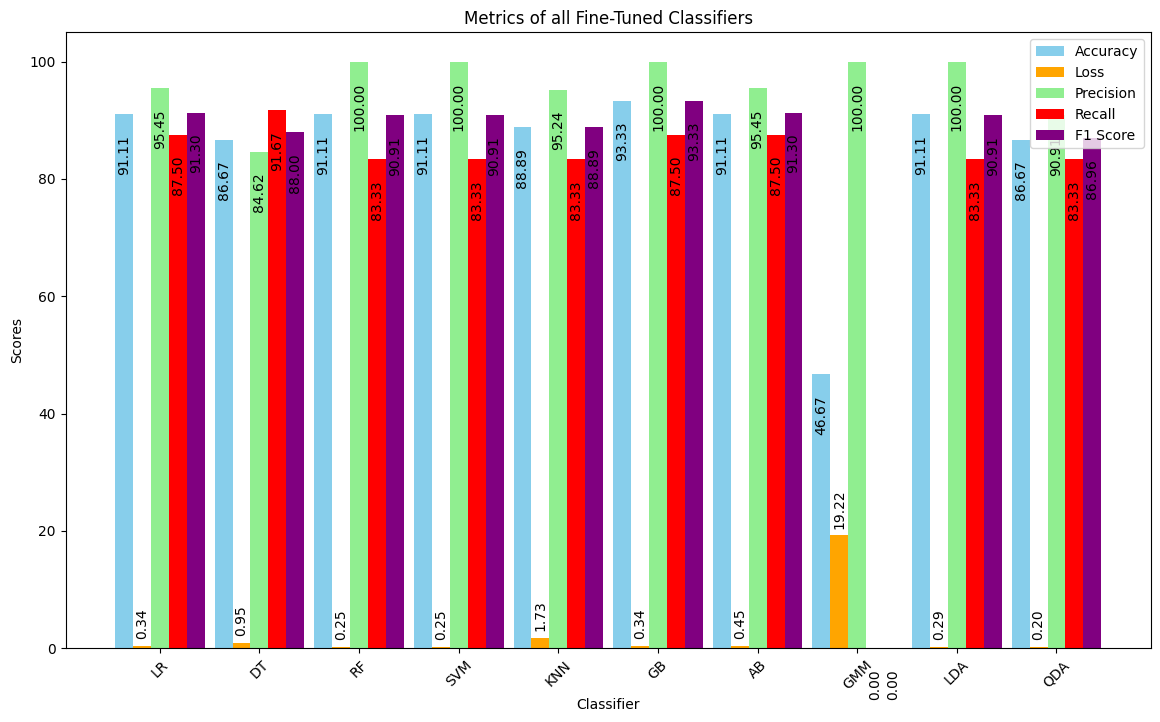}
     \caption{Training Phase Performance Metrics results for fine-tuned ML Models}
        \label{fig:figure1}
\end{figure*}

\begin{figure*}
    \centering
    \includegraphics[width=1\linewidth, height=0.35\textheight,keepaspectratio]{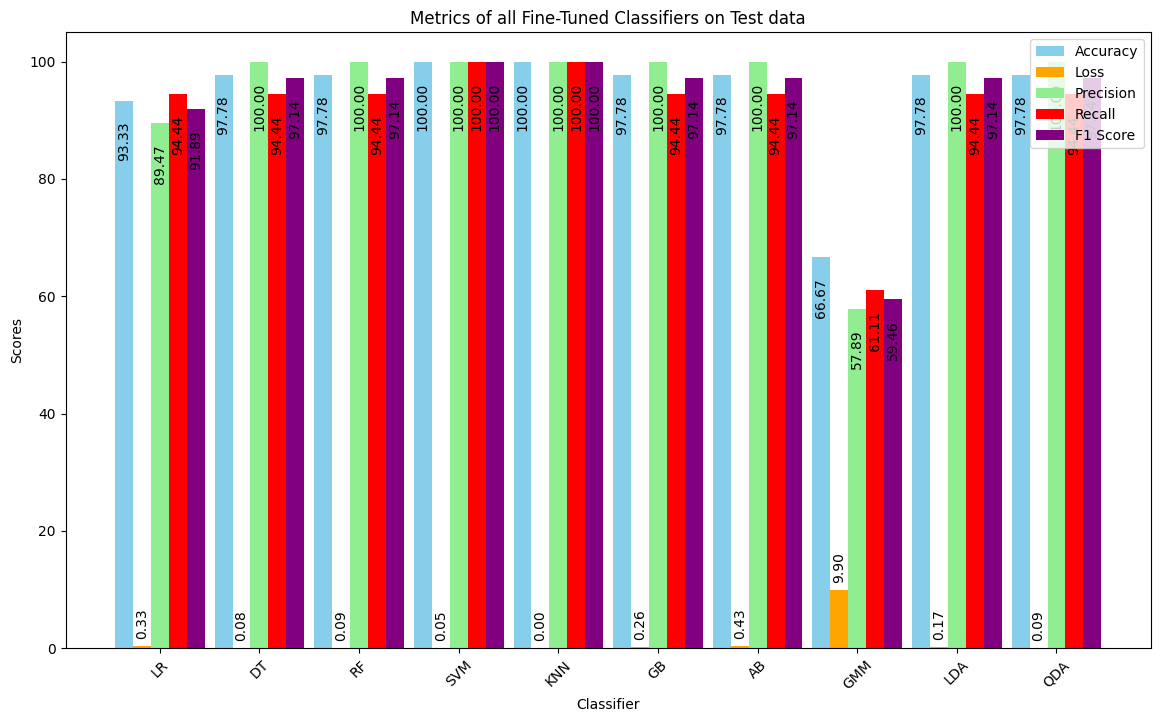} 
        \caption{Testing Phase Performance Metrics results for fine-tuned ML Models}
        \label{fig:figure2}
\end{figure*}

\begin{figure*}[!ht]
    \centering
    \includegraphics[width=0.18\textwidth]{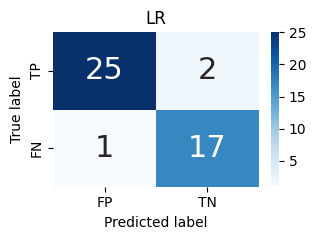}\hfill
    \includegraphics[width=0.18\textwidth]{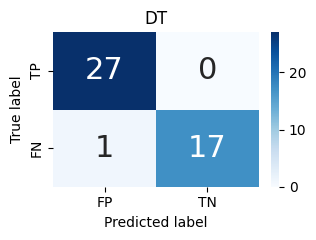}\hfill
    \includegraphics[width=0.18\textwidth]{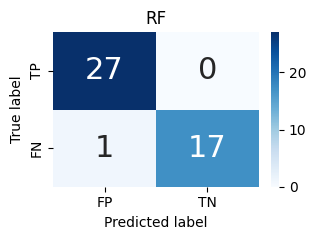}\hfill
    \includegraphics[width=0.18\textwidth]{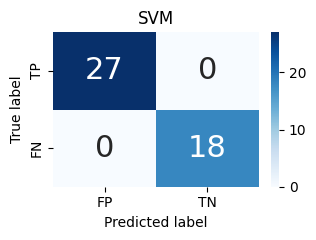}\hfill
    \includegraphics[width=0.18\textwidth]{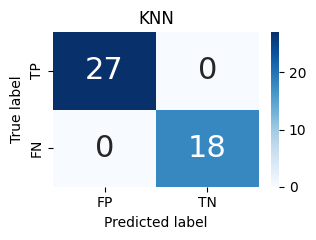}\\
    
    \includegraphics[width=0.18\textwidth]{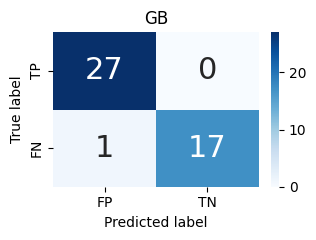}\hfill
    \includegraphics[width=0.18\textwidth]{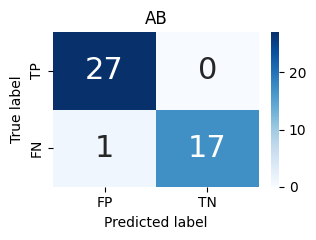}\hfill
    \includegraphics[width=0.18\textwidth]{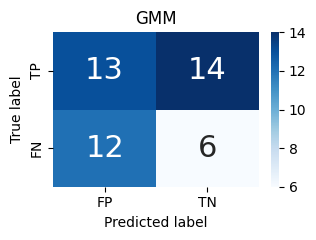}\hfill
    \includegraphics[width=0.18\textwidth]{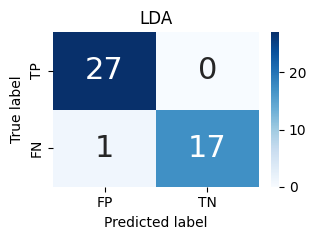}\hfill
    \includegraphics[width=0.18\textwidth]{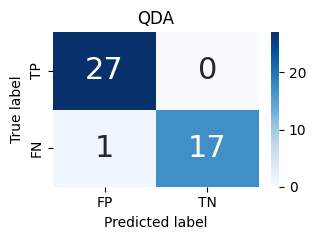}\\
    
    \caption{ML Model wise Confusion Matrix}
    \label{fig:cm}
\end{figure*}

\begin{figure}[ht]
    \centering
    \begin{minipage}{0.48\textwidth}
        \centering
        \includegraphics[width=\textwidth]{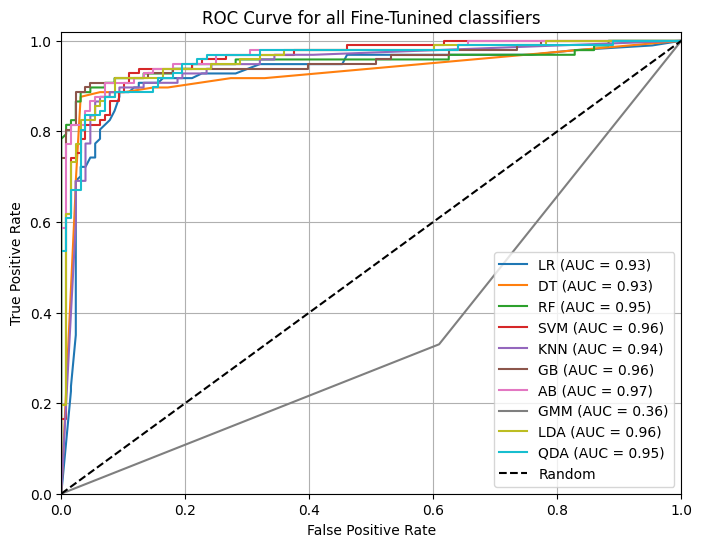}
        \caption{Fine-tuned ML Models AUC-ROC curve in training phase}
        \label{fig:enter-label1}
    \end{minipage}\hfill
    \begin{minipage}{0.48\textwidth}
        \centering
        \includegraphics[width=\textwidth]{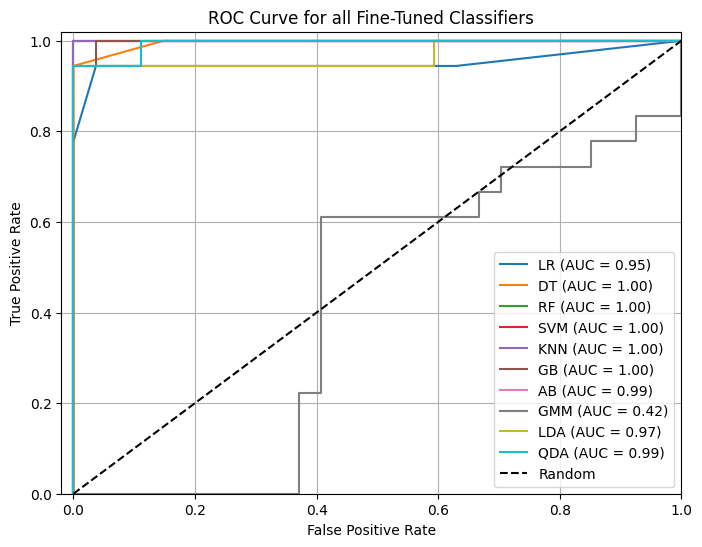}
        \caption{Fine-tuned ML Models AUC-ROC curve in testing phase}
        \label{fig:enter-label2}
    \end{minipage}
\end{figure}

We extensively evaluated all machine learning models using a detailed set of metrics. These metrics, detailed below, are strong indicators of the classifier's performance across multiple aspects.

\textbf{Accuracy:} This metric quantifies the ratio of observations the model correctly predicted out of all predictions made. It's the most straightforward way to gauge a model's performance, tallying up the true positives $T_p$ and true negatives $T_n$ against the overall predictions count. However, when the data is unevenly distributed across classes, this metric might not fully capture the model's efficacy.

{\footnotesize\[
    \text{Accuracy} = \frac{\text{Correctly predicted results}}{\text{Total predicted results}} = \frac{T_p + T_n}{T_p + T_n + F_p + F_n}
\]}

\textbf{Recall: } This index reflects the fraction of $T_p$ accurately pinpointed by the model out of all positive cases. It becomes essential when the consequences of overlooking an optimistic case (a false negative $F_n$) are significant. A model with a high sensitivity index can identify positive cases, yet it doesn't reveal the count of false positives $F_p$, which it might also predict.

\[
\text{Recall (RC)} = \frac{T_p}{T_p + F_n}
\]

\textbf{Log Loss: } It quantifies the accuracy of a probabilistic classification model, where each prediction lies on a spectrum from 0 to 1. It imposes a heavier penalty on predictions that are certain yet incorrect, considering the model's confidence relative to the actual class label. The ideal cross-entropy error is 0, indicating flawless predictions, while higher values suggest room for improvement.

\[
\text{Log Loss} = -\frac{1}{N} \sum_{i=1}^{N} [y_i \log(\hat{y_i}) + (1 - y_i) \log(1 - \hat{y_i})]
\]

\textbf{Positive Predictive Value:} Often termed Precision, this metric calculates the model's precision in identifying positive instances. It measures the model’s capacity to identify only pertinent cases. A model with a high Positive Predictive Value implies a minimal rate of $F_p$, though it doesn’t account for the occurrence of $F_n$. Precision becomes a critical factor in contexts where the implications of making a $F_p$ error are substantial.

\[
\text{Precision (PPV)} = \frac{T_p}{T_p + F_p}
\]

\textbf{Harmonic Mean of Precision and Recall:} Commonly known as the F1 Score, this measure combines PPV and RC into a single figure to evaluate their equilibrium. It's particularly beneficial for evaluating models on imbalanced datasets. The optimal F1 Score is 1, indicating impeccable PPV and RC, while the lowest possible score is 0.

\[
\text{F1 Score} = 2 \times \frac{\text{PPV} \times \text{RC}}{\text{PPV} + \text{RC}}
\]

\textbf{AUC-ROC Curve: } The AUC-ROC Curve is an evaluative metric for binary classifiers, reflecting their performance across varying thresholds. The ROC curve visualizes a classifier's capability by plotting the True Positive Rate (TPR) against the False Positive Rate (FPR) at multiple thresholds. The AUC measures how likely a model is to rank a positive sample above a negative one., with values spanning from 0 (entirely incorrect predictions) to 1 (flawless predictions).

The AUC-ROC offers a comprehensive metric for assessing a classifier's effectiveness across all possible thresholds, proving to be an essential tool in binary classification challenges. It excels by giving a singular performance indicator irrespective of the chosen cutoff, facilitating the comparison between different classifiers

\[
\text{TPR} = \frac{T_p}{T_p + F_n}
\]

\[
\text{FPR} = \frac{F_p}{F_p + T_n}
\]

\[
\text{AUC-ROC} = \int_{0}^{1} \text{TPR}(x) \, dx
\]

\subsection{PERFORMANCE ANALYSIS OF ALL FINE\_TUNED CLASSIFIERS WITH K = 25}
After the 5-fold cross-validation process, where we determined an optimal number of features (k = 25) for all ML models, we conducted a comprehensive performance analysis of all fine-tuned models under this parameter setting. The selection of k = 25 with 20 features is based on our earlier findings as discussed in Section \pageref{MV}, which highlighted this as the point beyond which no significant improvements in model performance were observed.


As Figure \ref{fig:figure1} vividly demonstrates, at the time of model training, LR, RF, SVM, GB, AB, and LDA all exhibited a commendable accuracy of 91\% - 93\%. However, when we delve into their precision values, we found RF, SVM, GB, and LDA outshining the rest with a perfect precision of 100\%. Moreover, based on other parameters, RF and SVM emerged as the clear winners, boasting the best performance with a minimum log-loss value of 0.25, a testament to their robustness and reliability. 

We further analyze RF and SVM based on their testing phase performance. As Figure \ref{fig:figure2} describes, SVM has the same precision as RF, but SVM has a 5.34\% higher recall value, a 2.22\% higher accuracy, and 0.02 less log loss than RF.

The SVM exhibited a slight recall advantage, indicating its strength in identifying the majority of positive cases as shown in Figure \ref{fig:cm}. This performance aspect is crucial in our applications where failing to detect true positives is significantly detrimental.

According to the AUC of training and testing 
as shown in Figure \ref{fig:enter-label1} and Figure \ref{fig:enter-label2}, the AdaBoost classifier displayed the highest level of discriminate capability, closely followed by SVM and RF. This suggests their potential for practical application in scenarios where a balance between $T_p$ and $F_p$ rates is crucial. These ROC and AUC results, along with the earlier analysis, reinforce the selection of the RF classifier for scenarios emphasizing the identification of positive cases, given its high recall and robust AUC value.

Considering all the performance parameters, we found that SVM performs best in its training and testing phase compared to other models with the best accuracy value of 100\%.
\section{DISCUSSION} \label{D}
\begin{table}[ht]
\centering
\caption{Comparison of Proposed and AMI tool statistics with DSM-5 results.}
\resizebox{\columnwidth}{!}
{
\begin{tabular}{lcc}
\toprule
& \multicolumn{2}{c}{\textbf{DSM-5 diagnosis}}\\
\cmidrule{2-3}
& \textbf{ASD (n = 27)} & \textbf{TD (n = 18)} \\
\midrule
\textbf{AMI Tool: ASD}      & 26  & 1   \\
\textbf{AMI Tool:  TD}      & 1   & 17  \\ \hline
\textbf{Proposed Tool: ASD} & 27  & 0   \\
\textbf{Proposed Tool:  TD} & 0   & 18  \\ 
\bottomrule
\end{tabular}
}
\label{Table4}
\end{table}

Comparative outcomes are summarized in a detailed Table \ref{Table4}, illustrating the diagnostic precision according to the DSM-5 criteria, the performance of our model, and AIIMS ASD tool \cite{gulati2019development}. Remarkably, our model demonstrated high accuracy in diagnosing ASD, aligning with the DSM-5 results. This was particularly notable given that our model achieved these results by utilizing a reduced set of 20 questions, fewer than the 28 questions employed by the AIIMS tool. This reduction in the number of questions without compromising diagnostic accuracy is a testament to the efficacy of our feature selection process and the robustness of our predictive algorithms. In Table \ref{Table4}, we can see the comparison study for both the AIIMS tool and our best model.

Further validating our model's performance, we subjected it to three different questionnaire examples provided in Gulati, S et al. \cite{Gulati} study. These examples served as real-world cases to assess our model's practical applicability. Impressively, our model accurately diagnosed all three examples, confirming its capability to replicate expert clinical judgment and suggesting its potential utility as a supportive diagnostic tool.

Our model's ability to deliver accurate diagnoses with fewer questions underscores the effectiveness of ML techniques in medical diagnostics and emphasizes the importance of precise feature selection. By identifying and focusing on the most informative questions, our approach offers a more efficient and potentially less burdensome alternative for initial screening processes.


\section{CONCLUSION} \label{C}
In conclusion, the integration of ML in healthcare, particularly in the early prediction of ASD, is demonstrating its potential to revolutionize patient assessment protocols. ASD is a rapidly growing neurodevelopmental disorder globally, and detecting it early is crucial for enhancing the life quality of those impacted. Traditional assessment methods, heavily reliant on human judgment, are not only time-intensive and costly but also less accessible to many. In light of this, our study embarked on creating a user-friendly, web-based ML solution to facilitate early ASD screening for the Indian population, available in both Hindi and English.

Utilizing real-world healthcare AMI data, our approach involved a meticulous selection process to identify the most indicative subset of screening questions. Twenty eight potential questions were analyzed using a majority voting feature selection method, ultimately determining that only twenty were necessary to maintain predictive accuracy with our ML model.

In the quest to select an optimal ML classifier for our application, different ML models were meticulously trained and fine-tuned across extensive hyperparameter configurations. The SVM classifier emerged as the most effective, exhibiting superior performance and reliability. It was, therefore, chosen to power our web-based solution, underscoring our commitment to providing a minimally invasive, efficient, and economical tool for preliminary ASD screening in the Indian context. This tool represents a significant stride toward making early ASD detection more accessible, facilitating timely intervention and support.

This research opens several avenues for further enhancement and refinement of ASD detection methods. While the current dataset comprising very limited patient responses has proven sufficient to train our models without overfitting, the scalability of our approach warrants caution. Working with a larger amount of real-time data in the future could make the model more robust. 

Moreover, the reliance on questionnaire responses, which are subject to the observations and interpretations of parents or caregivers, introduces a degree of subjectivity that could influence the screening outcomes. Acknowledging this limitation, future work will explore additional modalities that could complement or even enhance the predictive capabilities of our current system. Using computer vision and pattern recognition advancements, image and video-based analyses offer promising alternatives. Through the direct collection and analysis of behavioral indicators and physiological reactions, these methods could diminish dependence on indirect observations, offering a clearer and more detailed comprehension of ASD-related behaviors. Integrating these varied data streams has the potential to create more accurate and dependable tools for predicting ASD, setting the stage for refined diagnostic screenings and tailored treatment approaches.

\bibliographystyle{unsrtnat}
\bibliography{references}  






\end{document}